%% file: main.tex
\documentclass[conference]{IEEEtran}
\IEEEoverridecommandlockouts
\usepackage[T1]{fontenc}
\usepackage[utf8]{inputenc}

\usepackage{cite}
\usepackage{amsmath,amssymb,amsfonts}
\usepackage{graphicx}
\usepackage{textcomp}
\usepackage{xcolor}
\def\BibTeX{{\rm B\kern-.05em{\sc i\kern-.025em b}\kern-.08em
    T\kern-.1667em\lower.7ex\hbox{E}\kern-.125emX}}

\usepackage{booktabs}
\usepackage{algorithm}
\usepackage{mathrsfs}
\usepackage{colortbl}
\usepackage{multirow}
\usepackage{subcaption}
\usepackage{newfloat}
\usepackage{url}
\usepackage[hyperfootnotes=true,colorlinks,linkcolor=blue,anchorcolor=black,citecolor=blue,urlcolor=black]{hyperref}
\usepackage{float}
\usepackage{paralist}
\usepackage{stmaryrd}
\usepackage{pifont}
\usepackage{amsfonts}
\usepackage{multirow}
\usepackage{algorithmicx}
\usepackage{algpseudocodex} 
\usepackage{threeparttable}
\usepackage{soul}
\usepackage{xspace}

\usepackage{orcidlink}

\newtheorem{example}{Example}
\newtheorem{theorem}{Theorem}
\newtheorem{theorem*}{Theorem}
\newtheorem{definition}{Definition}
\newtheorem{problem}{Problem}

\newtheorem{proposition*}{Proposition}

\makeatletter
\newcommand{\linebreakand}{
\end{@IEEEauthorhalign}
\hfill\mbox{}\par\mbox{}\hfill\begin{@IEEEauthorhalign}
}
\makeatother

\allowdisplaybreaks[4]
\input{commands}

\begin{document}

\title{
% Model-Free Reinforcement Learning Guided by Causation Semantics of Signal Temporal Logic
% Causation-Guided Reinforcement Learning for Control Synthesis based Online Monitoring of Signal Temporal Logic
% Control Synthesis based Reinforcement Learning with Online Causation Monitoring of Signal Temporal Logic
% Online-Causation-Guided Reinforcement Learning for Control Synthesis of Cyber-Physical Systems
% Online-Causation-Guided Control Synthesis of Cyber-Physical\hspace{-0.1em} Systems\hspace{-0.1em} for\hspace{-0.1em} Real-Time\hspace{-0.1em} Specifications
Control Synthesis of Cyber-Physical Systems for Real-Time Specifications through Causation-Guided Reinforcement Learning
}

\author{

\IEEEauthorblockN{Xiaochen Tang\orcidlink{0000-0001-9895-6067}}
\IEEEauthorblockA{
\textit{School of Computer Science and Technology,} \\
\textit{Tongji University} \\
Shanghai, China \\
xiaochen9697@tongji.edu.cn} 
\and
\IEEEauthorblockN{Zhenya Zhang\textsuperscript{*}\orcidlink{0000-0002-3854-9846}}
\IEEEauthorblockA{
\textit{Faculty of Information Science and Electrical Engineering,} \\
\textit{Kyushu University, Fukuoka, Japan} \\
\textit{\& National Institute of Informatics, Tokyo, Japan}\\
zhang@ait.kyushu-u.ac.jp} 
\linebreakand
\IEEEauthorblockN{Miaomiao Zhang\textsuperscript{*}\orcidlink{0000-0001-9179-0893}}
\hspace{3cm}\IEEEauthorblockA{\textit{School of Computer Science and Technology,} \\
\textit{Tongji University}\\
Shanghai, China \\
miaomiao@tongji.edu.cn}
\and
\IEEEauthorblockN{Jie An\textsuperscript{*}\orcidlink{0000-0001-9260-9697}}
\IEEEauthorblockA{\textit{National Key Lab. of Space Integrated Information System,} \\
\textit{Institute of Software, Chinese Academy of Sciences,}\\
% \textit{\& University of Chinese Academy of Sciences} \\
Beijing, China \\
anjie@iscas.ac.cn}
\thanks{*~Corresponding authors.}
}

\maketitle

\begin{abstract}
In real-time and safety-critical cyber-physical systems (CPSs), control synthesis must guarantee that generated policies meet stringent timing and correctness requirements under uncertain and dynamic conditions. Signal temporal logic (STL) has emerged as a powerful formalism of expressing real-time constraints, with its semantics enabling quantitative assessment of system behavior. Meanwhile, reinforcement learning (RL) has become an important method for solving control synthesis problems in unknown environments. Recent studies incorporate STL-based reward functions into RL to automatically synthesize control policies. However, the automatically inferred rewards obtained by these methods represent the global assessment of a whole or partial path but do not accumulate the rewards of local changes accurately, so the sparse global rewards may lead to non-convergence and unstable training performances. In this paper, we propose an online reward generation method guided by the online causation monitoring of STL. Our approach continuously monitors system behavior against an STL specification at each control step, computing the quantitative distance toward satisfaction or violation and thereby producing rewards that reflect instantaneous state dynamics. Additionally, we provide a smooth approximation of the causation semantics to overcome the discontinuity of the causation semantics and make it differentiable for using deep-RL methods. We have implemented a prototype tool and evaluated it in the Gym environment on a variety of continuously controlled benchmarks. Experimental results show that our proposed STL-guided RL method with online causation semantics outperforms existing relevant STL-guided RL methods, providing a more robust and efficient reward generation framework for deep-RL.
\end{abstract}

\begin{IEEEkeywords}
control synthesis, CPS, STL, real-time specification, reinforcement learning
\end{IEEEkeywords}

\input{Chapter/1-introduction}
\input{Chapter/2-preliminaries}
\input{Chapter/3-problem_formulation}

\input{Chapter/4-experiments}

\input{Chapter/5-related_work}

\input{Chapter/6-conclusion}

\input{Chapter/8-acknowledgements}

\bibliographystyle{IEEEtran}
\bibliography{rtss25}

% \input{Chapter/10-appendix}
% \clearpage
% \input{Chapter/9-response}

\end{document}

%% file: commands.tex
\newcommand{\hlt}{\cellcolor{green!20}} % highlight table elements
\newcommand{\oomit}[1]{}

\newcommand{\bw}{\ensuremath{\mathbf{v}}\xspace}

\newcommand{\bwp}[2]{\ensuremath{\mathbf{v}_{#1:#2}}\xspace}
\newcommand{\Mon}[3]{\mathrm{[R]}(#1, #2, #3)}
\newcommand{\MonL}[3]{\mathrm{[R]}^\mathsf{L}(#1, #2, #3)}
\newcommand{\MonU}[3]{\mathrm{[R]}^\mathsf{U}(#1, #2, #3)}
\newcommand{\Rmax}{\mathtt{R}^{\alpha}_{\mathrm{max}}}
\newcommand{\Rmin}{\mathtt{R}^{\alpha}_{\mathrm{min}}}
\newcommand{\Robust}[3]{\rho(#1, #2, #3)}
\newcommand{\UntilOp}[1]{\mathbin{\mathcal{U}_{#1}}}

\newcommand{\InsRobVio}[3]{\mathcal{[\mathscr{R}]}^\ominus\left(#1, #2, #3\right)}

%% file: Chapter/1-introduction.tex
\section{Introduction}

{\em Cyber-physical systems} (CPSs)~\cite{ref_Baheti2011cyber,ref_Lee2008} are assigned increasingly complex objectives including reactive and sequential tasks in domains such as autonomous vehicles, advanced manufacturing, and health care systems. {\em Temporal logics}~\cite{conf/focs/Pnueli77} are widely adopted to specify properties and verify behaviors of CPSs due to their rich and rigorous expressiveness.
However, some tasks in the CPS have not only temporal requirements, but also more stringent deadline constraints.
{\em Signal Temporal Logic} (STL)~\cite{conf/formats/DonzeM10, journals/tcs/FainekosP09, journals/fmsd/JaksicBGNN18} introduces real-time intervals and continuous signal predicates to precisely express real-time constraints such as ``respond in $[a,b]$ time'' and ``continuously satisfy safety conditions''. 
The {\em robust semantics}~\cite{conf/formats/DonzeM10} of STL provides quantitative satisfaction measures that allow the system to assess the extent to which trajectories satisfy or violate the specification, while STL-based online monitors enable real-time monitoring and alerting of CPS.
In addition, the use of STL to guide the automatic synthesis of controllers has attracted research attention in CPS areas such as robotics~\cite{Jain2021} and transportation network control\cite{han2017privacy}.
Typical approaches are translating STL specifications into mixed-integer linear constraints and embedding them into model predictive control (MPC) frameworks~\cite{conf/cdc/RamanDMMSS14, journals/ras/BuyukkocakAY24}, and converting STL formulas into finite automata for automata-guided path planning~\cite{conf/amcc/LindemannD20, journals/ral/GundanaK21, conf/cdc/HoISL22}.
However, it is a great challenge to establish a reliable mathematical model due to the complex dynamics of the system and the intricate design objectives involved.

Model-free {\em reinforcement learning} (RL)~\cite{journals/tnn/SuttonB98} has emerged as an effective method to address complex control synthesis problems of CPSs with unknown environment dynamics. 
Recent works~\cite{conf/cdc/AksarayJKSB16} incorporate STL semantics into RL to generate rewards for complex RL tasks.
The powerful expressiveness of STL makes it effectively express complex control objectives in RL, which are difficult to achieve with a handcrafted reward function.
Moreover, the robust semantics of STL provides a quantitative way to show how much a trajectory satisfies or violates a system specification, which can significantly improve the stability and performance of control policy synthesis in complex CPS tasks using RL. 
Nevertheless, reward functions based on STL robust semantics are global, i.e., the rewards are only available over an entire episode, which will affect all states in a learning episode.
These non-Markovian reward functions may lead to non-convergence and unstable training performances. 
To mitigate the problem, within the deep-RL framework~\cite{ref_conf/iccps/GaoHZKZ19}, Balakrishnan and Deshmukh~\cite{conf/iros/0001D19} utilized the robustness value of a partial signal as the reward of its last state. 
Therefore, in each step, the reward of the current state can be computed based on the robustness value of the corresponding partial signal. 
Following the idea, Singh and Saha~\cite{conf/aaai/0004S23} proposed a new aggregation-based smooth approximation of STL semantics of partial signals additionally, making it more friendly to the existing deep-RL algorithms.

However, these STL-guided RL methods with single states as network inputs still suffer from the problem of non-Markovianity, since the reward functions are based on a partial sequence of states.
In addition, based on the robust semantics, it is difficult to solve the problem intrinsically, since the robust semantics of STL may not accurately describe the system's evolution at the current instant~\cite{conf/cav/SelyuninJNRHBNG17,journals/tcad/ZhangAX22}.
Recently, 
{\em online causation semantics}~\cite{conf/cav/ZhangAAH23} is proposed as an alternative of online robust semantics~\cite{journals/fmsd/DeshmukhDGJJS17} and provides more detailed information on changes of system states than the robust semantics. Unlike classic online robust monitoring approaches that aim to measure \emph{whether or not} a partial signal violates the given specification, causation semantics considers whether each instant can be treated as the \emph{causation} of the violation, thus offering a refined assessment of each system state against the specification. This feature of causation semantics coincidentally matches the spirit of RL rewards in control synthesis, which also aim to give a precise assessment for each system state, and therefore, this motivates us to explore the assignment of RL rewards based on causation semantics.

In this paper, we propose a novel online causation-guided RL method for control synthesis of real-time systems.
The main contributions of this paper are as follows.
\begin{itemize}
    \item[-] We applied the online causation monitor of STL to generate rewards for RL. 
    The technical adjustments performed to make the online causation monitor suitable for RL involve: 
    \begin{itemize}
        \item[.] $\tau$-MDP~\cite{conf/cdc/AksarayJKSB16} is adopted to accommodate non-Markovianity;
        \item[.] Sampling windows are proposed to avoid dealing with long trajectories that can diminish efficiency in RL;
        \item[.] Smoothing approximation of the online causation semantics to meet the differentiable requirement of deep-RL.
    \end{itemize}
    \item[-] We have implemented a prototype and evaluated it on several challenging continuous control benchmarks available in the Gym environment~\cite{ref_journals/corr/BrockmanCPSSTZ16}.
\end{itemize}

We compare our method with the baseline RL method using hand-crafted reward functions and the existing STL-guided RL method using the robust semantics. 
Experimental results demonstrate that our method converges more stably than other methods, and the trained control policies achieve higher values on three evaluation metrics.

%% file: Chapter/2-preliminaries.tex
\section{Preliminaries} \label{sec_preliminaries}

\subsection{Signal Temporal Logic} \label{subsec_STL}
STL is a widely adopted formal language that can describe the real-time properties of a system.
Let $T\in\mathbb{R}_+$ be a positive real, and $d\in\mathbb{N}_+$ be a positive integer. 
A \textit{d-dimensional signal} is a function $\bold{v}:[0,T]\rightarrow \mathbb{R}^d$, where $T$ is called the time {\em horizon} of $\bold{v}$.
Given an arbitrary time instant $t\in[0, T ]$, $\bold{v}(t)$ is a $d$-dimensional real vector,
each dimension concerning a signal variable that has a certain physical meaning. 
In this paper, we ﬁx a set $\bold{Var}$ of variables and assume that a signal $\bold{v}$ is spatially bounded, i.e., for all $t\in[0,T]$, $\bold{v}(t)\in\Omega$, where $\Omega$ is a $d$-dimensional hyper-rectangle. 
The syntax of STL is defined as follows.

\begin{definition}[STL syntax] \label{def_stl}
In STL, the atomic propositions $\alpha$ and the formulas $\varphi$ are defined as follows: 
\begin{equation}
\begin{aligned}
    &\alpha::\equiv f(x_1,...,x_K)>0&\\
    &\varphi::\equiv \alpha \mid \top \mid \neg\varphi \mid \varphi_1\wedge\varphi_2 \mid \Box_{I}\varphi \mid \Diamond_{I}\varphi \mid \varphi_1 \mathcal{U}_I \varphi_2&
\end{aligned}
\end{equation}
where $\top$ means $true$, $f$ is a $K$-ary function $f:\mathbb{R}^K\rightarrow\mathbb{R}$, $x_1,...,x_K\in\bold{Var}$, and $I$ is a closed interval over $\mathbb{R}_+$, i.e., $I=[l, u]$, where $l,u \in \mathbb{R}_+$ and $l\le u$.
The operators $\neg$ and $\wedge$ denote logical NOT and AND operators.
Other logical operators like logical OR ($\vee$) and implication ($\Rightarrow$) can be derived using $\neg$ and $\wedge$, where $\varphi_1\vee\varphi_2\equiv\neg(\neg\varphi_1\wedge\neg\varphi_2)$ and $\varphi_1\Rightarrow\varphi_2\equiv\neg\varphi_1\vee\varphi_2$.
The temporal operator $\mathcal{U}_I$ is the Until operator implying that $\varphi_2$ becomes true sometime in the time interval $I$ and $\varphi_1$ must remain true until $\varphi_2$ becomes true. 
The other two temporal operators are known as Eventually ($\Diamond_I$) and Always ($\Box_I$), where $\Diamond_I\varphi\equiv\top \mathcal{U}_I\varphi$ and $\Box_I\varphi\equiv\neg\Diamond_I\neg\varphi$.
\end{definition}
\begin{definition}[STL robust semantics]\label{def_stl_robust}
    Let $\bold{v}$ be a signal, $\varphi$ be an STL formula and $\mu\in \mathbb{R}_+$ be an instant. 
    The $robustness$ $\rho(\bold{v},\varphi,\mu)\in\mathbb{R}\cup\{\infty,-\infty\}$ of $\bold{v}$ w.r.t. $\varphi$ at $\mu$ is defined by induction on the construction of formulas, as:
    \begin{align*}
        &\rho(\bold{v},\alpha,\mu):=f(\bold{v}(\mu)),&\\        
        &\rho(\bold{v},\top,\mu):=\infty,&\\
        &\rho(\bold{v},\neg\varphi,\mu):=-\rho(\bold{v},\varphi,\mu),& \\        &\rho(\bold{v},\varphi_1\wedge\varphi_2,\mu):=\min(\rho(\bold{v},\varphi_1,\mu),\rho(\bold{v},\varphi_2,\mu)),&\\
        &\rho(\bold{v},\Box_I\varphi,\mu):=\inf_{t\in\mu+I}\rho(\bold{v},\varphi,t),&\\
        &\rho(\bold{v},\Diamond_I\varphi,\mu):=\sup_{t\in\mu+I}\rho(\bold{v},\varphi,t),&\\       &\rho(\bold{v},\varphi_1 \mathcal{U}_I\varphi_2,\mu) := \sup_{t\in\mu+I}\min(\rho(\bold{v},\varphi_2,t),\inf_{t'\in[\mu,t)}\rho(\bold{v},\varphi_1,t')).&
    \end{align*}
    Here, $\mu+I$ denotes the interval $[l+\mu,u+\mu]$.
\end{definition}

The robustness of a signal $\bold{v}$ w.r.t a specification $\varphi$ is defined as $\rho(\bold{v},\varphi)= \rho(\bold{v},\varphi,0)$, i.e., the robustness at time $0$.
The original STL semantics is Boolean, which represents whether a signal $\bold{v}$ satisfies $\varphi$ at an instant $\mu$, i.e., $(\bold{v},\mu)\vDash\varphi$. 
The robust semantics in Definition~\ref{def_stl_robust} is a quantitative extension that refines the original Boolean STL semantics, in the sense that, $\rho(\bold{v},\varphi,\mu)>0$ implies $(\bold{v},\mu)\vDash\varphi$ and $\rho(\bold{v},\varphi,\mu)<0$ implies $(\bold{v},\mu)\nvDash\varphi$.
% It quantifies how well a given signal $\bold{v}$ satisfies a given formula $\varphi$. 
Intuitively, it quantifies how strongly a given signal $\bold{v}$ satisfies or violates a given formula $\varphi$.
% the robustness of a signal $\bold{v}$ quantifies how strongly a system trajectory satisfies or violates an STL specification.

For some discrete systems, control synthesis with STL specifications is decidable and tractable. For example, \cite{conf/cdc/RamanDMMSS14} formulates the STL control synthesis problem as a mixed integer linear program.
However, in general, especially for complex continuous systems, the STL control problem is undecidable or computationally intractable.

% Add: classic online monitor of STL
\subsection{Online Robust Semantics of STL}
STL robust semantics in Definition~\ref{def_stl_robust} provide an offline monitoring approach for {\em complete signals}. {\em Online monitoring}, instead, targets a growing {\em partial signal} at runtime.
% Besides the verdicts $\top$ and $\bot$, an online monitor can also report the verdict \textunknown (denoted as $\unknown$), which represents a status when the satisfaction of the signal to $\varphi$ is not decided yet. 
The formal definitions of partial signals and the online robust semantics of STL are as follows.
% In the following, we formally define partial signals and introduce online monitors for STL.

For a signal $\bw$ with the time horizon $T$, 
% Let $T$ be the time horizon of a signal $\bw$, and 
let $[a, b]\subseteq [0, T]$ be a sub-interval in the time domain $[0, T]$. 
A {\em partial signal} $\bwp{a}{b}$ is a function which is only defined in the interval $[a, b]$; in the remaining domain $[0,T]\setminus[a, b]$, we denote the value of $\bwp{a}{b}$ at those instants as $\epsilon$, where $\epsilon$ stands for a value that is not defined. 
Specifically, if $a = 0$ and $b\in(a, T]$, a partial signal $\bwp{a}{b}$ is called a {\em prefix} (partial) signal; dually, if $b = T$ and $a\in[0, b)$, a partial signal $\bwp{a}{b}$ is called a {\em suffix} (partial) signal.
Given a prefix signal \bwp{0}{b}, a {\em completion} $\bwp{0}{b}\cdot\bwp{b}{T}$ of \bwp{0}{b} is defined as the concatenation of \bwp{0}{b} with a suffix signal \bwp{b}{T}.

\begin{definition}[Online STL robust semantics]\label{def_classic_Monitor}
    Let $\bwp{0}{b}$ be a prefix signal, and let $\varphi$ be an STL formula. We denote by $\Rmax$ and $\Rmin$ the possible {\em maximum} and {\em minimum bounds} of the robustness $\Robust{\bw}{\alpha}{\mu}$\footnote{$\Robust{\bw}{\alpha}{\mu}$ is bounded because $\bw$ is bounded by $\Omega$. In practice, if $\Omega$ is not know, we set $\Rmax$ and $\Rmin$ to, respectively, $\infty$ and $-\infty$.}. Then, an online robust semantics $\Mon{\bwp{0}{b}}{\varphi}{\mu}$, which returns a sub-interval of $[\Rmin, \Rmax]$ at the instant $b$, is defined as follows, by induction on the construction of formulas.
    \begin{align*}
        &\Mon{\bwp{0}{b}}{\alpha}{\mu} := \begin{cases}
 \big[f\left(\bwp{0}{b}(\mu)\right), f\left(\bwp{0}{b}(\mu)\right)\big] & \text{if } \mu\in [0, b]\\
 \big[\Rmin, \Rmax\big]& \text{otherwise}
 \end{cases}\\
 &\Mon{\bwp{0}{b}}{\neg\varphi}{\mu} := -\Mon{\bwp{0}{b}}{\varphi}{\mu} \\
 &\Mon{\bwp{0}{b}}{\varphi_1\land\varphi_2}{\mu} := \\ &\qquad\min\Big(\Mon{\bwp{0}{b}}{\varphi_1}{\mu}, \Mon{\bwp{0}{b}}{\varphi_2}{\mu}\Big) \\
 &\Mon{\bwp{0}{b}}{\Box_I\varphi}{\mu} := \inf_{t\in \mu+I}\Big(\Mon{\bwp{0}{b}}{\varphi}{t}\Big) \\
 &\Mon{\bwp{0}{b}}{\varphi_1\UntilOp{I}\varphi_2}{\mu} :=\\ &\qquad\sup_{t\in\mu +I}\min\Big(\Mon{\bwp{0}{b}}{\varphi_2}{t}, \inf_{t'\in[\mu, t)}\Mon{\bwp{0}{b}}{\varphi_1}{t'}\Big)
    \end{align*}
  Here, $f$ is defined as in Definition~\ref{def_stl}, and the arithmetic rules over intervals $I=[l, u]$ are defined as follows:
  %\begin{align*}
      $-I:=[-u, -l] \text{ and } \min(I_1,I_2) := [\min(l_1, l_2), \min(u_1, u_2)]$.
  %\end{align*}
\end{definition}
We denote $\MonU{\bwp{0}{b}}{\varphi}{\mu}$ and $\MonL{\bwp{0}{b}}{\varphi}{\mu}$ as the upper bound and the lower bound of $\Mon{\bwp{0}{b}}{\varphi}{\mu}$ respectively. Intuitively, the two bounds together form the reachable robustness interval of the completion $\bwp{0}{b}\cdot\bwp{b}{T}$, under any possible suffix signal \bwp{b}{T}. 
% For instance, in Fig.~\ref{fig_example}, the upper bound $\MonUnoArgs$ at $b= 20$ is 0, which indicates that the robustness of the completion of the signal \speed, under any suffix, can never be larger than 0.

\subsection{Markov Decision Processes} \label{subsec_MDP}
A {\em Markov decision process (MDP)}~\cite{ref_Bellman1957} is a transition system in which state transitions take place non-deterministically following a probability distribution. 
\begin{definition} [MDP]
    An MDP is a tuple $\mathcal{M}=\langle S, A, P, R \rangle$, where 
    \begin{itemize}
    \item[-] $S$ is a set of states and denotes the state-space, 
    \item[-] $A$ is a finite set of input actions, 
    \item[-] $P:S\times A\times S \rightarrow [0,1]$ is the probabilistic transition function where $P(s,a,s')$ denotes the probability of the transition from state $s$ to $s'$ by action $a$, 
    \item[-] $R: S\rightarrow \mathbb{R}$ is a real-valued reward function.
    \end{itemize}
\end{definition}
A sequence of states is called a {\em trajectory}.
For any integer $i,j\in \mathbb{N}_{\geq 0}$, let $s_i\in S$ denote the $i$-th state of an arbitrary trajectory in system ($s_0$ is the initial state), and $s_{i:j}$ denote the partial trajectory from the $i$-th state to the $j$-th state. Note that $s_{i:i} = s_i$.
Let $\omega$ denote a finite-state trajectory, $\omega(i)$ denote the $i$-th element of $\omega$, $\omega(-1)$ denote the last element of $\omega$, and $\omega(i,j)$ denote the sub-trajectory of $\omega$ from the $i$-th element to the $j$-th element.
A {\em policy} $\pi: S\rightarrow Dist(A)$ for a system $\mathcal{M}$ is a function that maps each state to a distribution over the actions.

A policy $\pi$ on $\mathcal{M}$ can induce a {\em discrete-time Markov chain (DTMC)} $\mathcal{D}_\mathcal{M}^{\pi}=\left\langle \bar{S}, \bar{P} \right\rangle$, where $\bar{S}=S$ is a set of states and $\bar{P}:\bar{S}\rightarrow Dist(\bar{S})$ is a probabilistic transition function that maps a current state to a distribution over the next states, and $\bar{P}(\bar{s}, \bar{s}') = \sum_{a\in A}\pi(a|\bar{s})P(\bar{s}, a, \bar{s}')$ for any $\bar{s},\bar{s}' \in \bar{S}$.

\subsection{STL-Guided RL Methods}

In control policy synthesis problems for complex CPSs with unknown models, STL-guided RL methods model the system models as MDPs, specify the system tasks in terms of STL specifications, and employ STL semantics to design the reward functions for RL.
In these methods, the control policy synthesis problem for an STL specification can be formalized as follows.
\begin{problem}[Control policy synthesis problem of STL semantics]\label{prob_synthesis}
    Let $\varphi$ be an STL specification that describes a task.
    Given an MDP model of the system $\mathcal{M}=\langle S, A, P, R_{\rho} \rangle$, where $P$ is unknown and the reward $R_{\rho}$ is designated by a specific robust semantics $\tilde{\rho}$. 
    The objective is to synthesize a control policy $\pi^*$, such that the expected robustness values of all state trajectories $\omega$ in $\mathcal{D}^{\pi^*}_{\mathcal{M}}$ w.r.t. $\varphi$ gets maximized.
    Mathematically,
    \begin{align}
        \pi^*=\mathop{\arg\max}_{\pi}\mathop{\mathbb{E}}_{\omega\sim \mathcal{D}_{\mathcal{M}}^{\pi}}[\tilde{\rho}(\omega, \varphi, 0)].
    \end{align}
\end{problem}
Note that $R_{\rho}$ can be defined over multiple STL semantics, yielding different reward functions.

As we described in the introduction, robust semantics and its derived approximation semantics have already been explored to solve the above control problem~\cite{conf/cdc/AksarayJKSB16, conf/iros/0001D19, conf/aaai/0004S23}. 
However, the robust semantics leads to the information masking problem~\cite{conf/rv/CimattiTT1,conf/cav/SelyuninJNRHBNG17,journals/tcad/ZhangAX22,conf/cav/ZhangAAH23}, which means that it may fail to report new violations or accurately describe the system's evolution at the current instant after the first violation. When robust semantics is applied in RL, the masked information leads to sparse, inaccurate, and non-ideal rewards, which will degrade the performance of RL and hinder policy convergence. Therefore, existing RL methods based on STL robust semantics may suffer from the information masking problem. 

%% file: Chapter/3-problem_formulation.tex
\section{Online Causation-Guided RL Method} \label{sec_problem}

In order to solve the problems of existing STL-guided RL methods, we propose the causation-guided RL method, which mainly consists of 1) generating RL rewards with the online causation semantics of STL in order to solve the problem of information masking of the existing STL's robust semantics, and 2) defining the environment model as $\tau$-MDP in order to solve the problem of non-Markovianity.

\subsection{Online Causation Semantics of STL} \label{subsec_cauSTL}

To solve the information masking problem, \cite{conf/cav/ZhangAAH23} proposed the STL causation semantics, which is able to provide more information on the system evolution than the STL robust semantics. 
Given a partial signal $\bold{v}_{0:b}$ from time $0$ to the current instant $b$, the causation semantics calculates a \textit{violation causation distance} $\mathscr{[R]}^{\ominus}$ and a \textit{satisfaction causation distance} $\mathscr{[R]}^{\oplus}$, which, respectively, indicate how far the signal value (the trajectory state) at
the current instant $b$ is from turning $b$ into a violation/satisfaction causation instant such that $b$ is relevant to the violation/satisfaction of a given STL formula. 

In this paper, since we only use the violation causation distance (explained in Section~\ref{subsec_target_spec}), we present the causation semantics as follows by omitting the satisfaction causation distance. The full definition can be found in~\cite{conf/cav/ZhangAAH23}.

\begin{definition}[Online causation semantics of STL]
\label{def_causation}
Let $\bold{v}_{0:b}$ be a partial signal from the initial instant $0$ to the current instant $b$, and $\varphi$ be an STL formula. 
At instant $b$, the violation causation distance $\mathscr{[R]}^{\ominus}(\bold{v}_{0:b},\varphi,\mu)$ can be computed as follows:
\begin{align*}
    &\mathscr{[R]}^{\ominus}(\bold{v}_{0:b},\alpha,\mu) := \begin{cases}
        f(\bold{v}_{0:b}(\mu)) & \text{if}\quad b=\mu\\
        \mathtt{R}^{\alpha}_{\mathrm{max}} & \text{otherwise}
    \end{cases}\\
    &\mathscr{[R]}^{\ominus}(\bold{v}_{0:b},\neg\varphi,\mu):= -\mathscr{[R]}^{\oplus }(\bold{v}_{0:b},\varphi,\mu)\\
    &\mathscr{[R]}^{\ominus}(\bold{v}_{0:b},\varphi_1\wedge\varphi_2, \mu):=\\ &\quad\min\left(\mathscr{[R]}^{\ominus}(\bold{v}_{0:b},\varphi_1, \mu), \mathscr{[R]}^{\ominus}(\bold{v}_{0:b},\varphi_2, \mu)\right) \\
    &\mathscr{[R]}^{\ominus}(\bold{v}_{0:b},\varphi_1\vee\varphi_2, \mu):= \\
    &\quad \min\left(
    \begin{array}{l}
         \max\left(
         \mathscr{[R]}^{\ominus}(\bold{v}_{0:b},\varphi_1, \mu), \mathrm{[R]}^{\mathsf{U}}(\bold{v}_{0:b},\varphi_2, \mu)
         \right),  \\
        \max\left(
        \mathrm{[R]}^{\mathsf{U}}(\bold{v}_{0:b},\varphi_1, \mu), \mathscr{[R]}^{\ominus}(\bold{v}_{0:b},\varphi_2, \mu)
        \right)
    \end{array}
    \right)\\
    &\mathscr{[R]}^{\ominus}(\bold{v}_{0:b},\Box_I\varphi, \mu):= \inf_{t\in\mu+I}\left(\mathscr{[R]}^{\ominus}(\bold{v}_{0:b},\varphi, t)\right)\\
    &\mathscr{[R]}^{\ominus}(\bold{v}_{0:b},\Diamond_I\varphi, \mu):=\\
    &\quad\inf_{t\in\mu+I}\left(
    \max\left(
    \mathscr{[R]}^{\ominus}(\bold{v}_{0:b},\varphi, t), \mathrm{[R]}^{\mathsf{U}}(\bold{v}_{0:b},\Diamond_I\varphi_2,\mu)
    \right)
    \right)\\
    &\mathscr{[R]}^{\ominus}(\bold{v}_{0:b},\varphi_1 \mathcal{U}_I \varphi_2, \mu):= \\
    & \inf_{t\in\mu+I}\left(
    \max\left(
    \begin{array}{l}
         \min\left(
         \begin{array}{l}
              \displaystyle\inf_{t'\in[\mu,t)}\mathscr{[R]}^{\ominus}(\bold{v}_{0:b},\varphi_1, t'),  \\
              \mathscr{[R]}^{\ominus}(\bold{v}_{0:b},\varphi_2, t)
         \end{array}
         \right),\\
          \mathrm{[R]}^{\mathsf{U}}(\bold{v}_{0:b},\varphi_1 \mathcal{U}_I \varphi_2, \mu)
    \end{array}
    \right)
    \right).
\end{align*}
% Here, $\mathtt{R}^{\alpha}_{\mathrm{max}}$ is the possible \textit{maximum bounds} of the robustness $\rho(\bold{v}_{0:b},\varphi, \mu)$. $\mathrm{[R]}^\mathsf{U}$, coming from~\cite{journals/fmsd/DeshmukhDGJJS17}, represents the maximal reachable robustness of any possible suffix signal (i.e., any continuation of the system's evolution)~\footnote{$\mathscr{[R]}^{\ominus}$ is bounded because $\bold{v}$ is bounded by $\Omega$. 
% In practice, if $\Omega$ is not know, we set $\mathtt{R}^{\alpha}_{\mathrm{max}}$ to $\infty$. 
% }. 
\end{definition}

\begin{figure}[t]
    \begin{subfigure}[b]{0.98\linewidth}
    \includegraphics[width=0.93\linewidth]{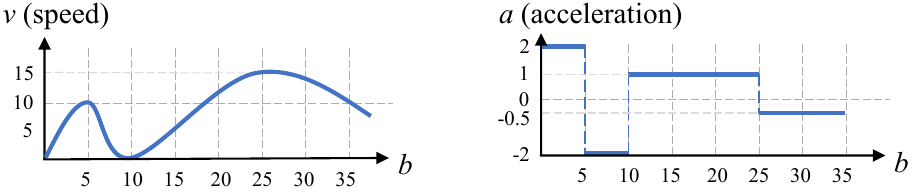}
    \caption{Example signals}
    \label{fig_example}
    \end{subfigure}\\
    \begin{subfigure}[b]{0.98\linewidth}
        \includegraphics[width=0.98\linewidth]{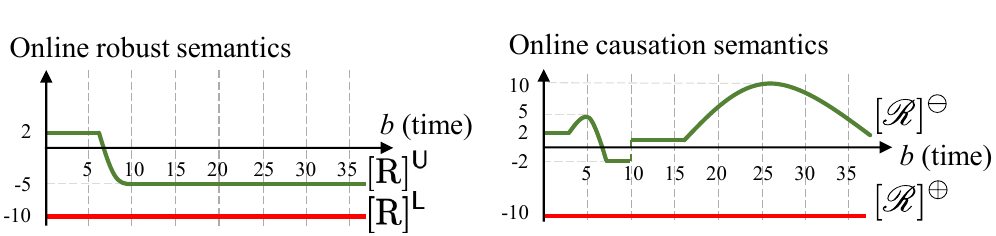}
    \caption{The illustrations for the corresponding online causation semantics and robust semantics}
    \label{fig_semantics}
    \end{subfigure}
    \caption{The comparison of online causation semantics and robust semantics against STL specification $\varphi = \Box_{[0,100]}(v>5 \vee a>0)$.}
    \label{fig_QCauM}
\end{figure}

Intuitively, violation causation distance $\mathscr{[R]}^{\ominus}(\bold{v}_{0:b},\varphi,\mu)$ represents the spatial distance of the signal value $\bold{v}_{0:b}$, at the current instant $b$, from turning $b$ into a violation causation instant such that $b$ is relevant to the violation of $\varphi$ (also applied to the satisfaction case dually). In this way, the causation semantics reflects the instantaneous status of system evolution more faithfully than robust semantics. 
Here is a simple example of computing the satisfaction causation distance.
\begin{example}
    Figure~\ref{fig_example} presents a vehicle's speed and acceleration signals.
    The specification requires that the vehicle always accelerates or its speed is always greater than $5$. For convenience, we name the sub-formulas of $\varphi$ as follows:
    \begin{gather*}
    \alpha_1\equiv v>5, \quad \alpha_2\equiv a>0, \quad
    f_1 \equiv v-5, \quad f_2\equiv a
    % \alpha_1 \equiv v >5 \qquad \alpha_2 \equiv a< 0
    \end{gather*}    
    At $b = 25$, the violation causation distance is computed as:
% {\mathcompact
\footnotesize
\begin{align*}
    &\InsRobVio{\bwp{0}{5}}{\varphi}{0} = \inf_{t\in[0,100]}\InsRobVio{\bwp{0}{5}}{\alpha_1 \vee \alpha_2}{t} \\
    =&\inf_{t\in[0,100]}\left(\min\left(
    \begin{array}{l}
         \max\left(
            \InsRobVio{\bwp{0}{5}}{\alpha_1}{t},  
            \MonU{\bwp{0}{5}}{\alpha_2}{t}
         \right), \\
          \max\left(
               \MonU{\bwp{0}{5}}{\alpha_1}{t},  
                \InsRobVio{\bwp{0}{5}}{\alpha_2}{t}
          \right)    
    \end{array}
    \right)\right) \\
    =&\min\left(
    \begin{array}{l}
         \max\left(
            \InsRobVio{\bwp{0}{5}}{\alpha_1}{5},  
            \MonU{\bwp{0}{5}}{\alpha_2}{5}
         \right), \\
          \max\left(
               \MonU{\bwp{0}{5}}{\alpha_1}{5},  
                \InsRobVio{\bwp{0}{5}}{\alpha_2}{5}
          \right)    
    \end{array}
    \right)\\
    =&\min\left(
    \begin{array}{l}
         \max\left(f_1(10), f_2(-2) \right), \\
         \max\left(f_1(10), f_2(-2) \right)    
    \end{array}
    \right)\\
    % =& \max\left(\MonU{\bwp{0}{5}}{\alpha_1}{5}, \MonU{\bwp{0}{5}}{\alpha_2}{5}\right)\\
    =& \max\left(f_1(10), f_2(-2) \right)=\max(5, -2) = 5
\end{align*}
% }
\normalsize
Figure~\ref{fig_semantics} shows a comparison between the corresponding online robust semantics and the causation semantics. 
\end{example}

Intuitively, in the interval $[20, 25]$, the system should satisfy the specification ``more'', as the vehicle accelerates and the speed increases. We can see the causation semantics reflects the system evaluation as $\mathscr{[R]}^{\ominus}$ increases in $[20, 25]$. However, robust semantic $\mathrm{[R]}^{\mathsf{U}}$ is monotonic and masks that as shown in the figure.

% For instance, Figure~\ref{fig_example} presents a vehicle's speed and acceleration signals. 
% For STL specification $\varphi = \Box_{[0,100]}(v>10 \Rightarrow \Diamond_{[0,5]}(a<0))$, Figure~\ref{fig_semantics} shows a comparison between the corresponding online robust semantics and the causation semantics. 
% For STL specification $\varphi = \Box_{[0,100]}(v>5\vee a>0)$, 

For the sake of simplicity, we use $\bar{\rho}$ to uniformly denote the violation causation distance or the satisfaction causation distance in the rest of this paper.
Note that the causation concept is not in the scope of the well-known causality theory~\cite{pearl2009causality}. More details can be found in~\cite{conf/cav/ZhangAAH23}.

\subsection{System Model} \label{subsec_system_model}
In regular RL problems, the system is usually modeled as an MDP.
However, in STL-guided RL methods, rewards are given based on a state trajectory rather than a single state.
Thus, as noted in~\cite{conf/cdc/AksarayJKSB16}, the RL algorithm cannot be applied to the original MDP, as the reward function is non-Markovian. 
To address this issue, we model the system as the $\tau$-MDP proposed in~\cite{conf/cdc/AksarayJKSB16}, which aggregates the state trajectories of the original MDP into new states.

\begin{definition} [$\tau$-MDP]
    Given an MDP $\mathcal{M}=\langle S, A, P, R \rangle$ and a positive integer $k \geq 2$ as state aggregation value,
    a $\tau$-MDP is a tuple $\mathcal{M}^{\tau}=\langle S^{\tau}, A, P^{\tau}, R^{\tau} \rangle$, where 
    \begin{itemize}
    \item[-] $S^{\tau}\subseteq (S)^{k}$ is a set of states, 
    Each state $s^{\tau} \in S^{\tau}$ corresponds to a $k$-length trajectory on $S$. 
    For an integer $t\in \mathbb{N}$, if $t\geq k-1$, $s_{t}^{\tau} = s_{t-k+1:t}$, otherwise $s_{t}^{\tau} = (s_0)^{k-t-1}\cdot s_{0:t}$;
    \item[-] $A$ is a finite set of input actions, which is the same as that in $\mathcal{M}$;
    \item[-] $P^{\tau}:S^{\tau}\times A\times S^{\tau} \rightarrow [0,1]$ is the probabilistic transition function. For any $s_{t}^{\tau}, s_{t+1}^{\tau}\in S^{\tau}$ and $a_t\in A$, $P^{\tau}(s_t^{\tau}, a, s_{t+1}^{\tau})=P(s_t,a,s_{t+1})$ if $s_{t}^{\tau}(1,k-1)=s_{t+1}^{\tau}(0,k-2)$, $P^{\tau}(s_t^{\tau}, a, s_{t+1}^{\tau})=0$ otherwise, 
    \item[-] $R^{\tau}: S^{\tau}\rightarrow \mathbb{R}$ is a real-valued reward function.
    \end{itemize}
\end{definition}
Intuitively, $\tau$-MDP treats the final $k$ step trajectory as an entire state, where $k$ is determined case by case.
The trajectory of a $\tau$-MDP is denoted as $\hat{\omega}$, each element $\hat{\omega}(i)$ is a $k$-length state trajectory.
Accordingly, we denote the policy as $\hat{\pi}: S^{\tau}\rightarrow Dist(A)$ and the DTMC as $\mathcal{D}_{\mathcal{M}^{\tau}}^{\hat{\pi}}=\left\langle \bar{S}^{\tau}, \bar{P}^{\tau} \right\rangle$.

Furthermore, there exists a one-to-one correspondence between the trajectories in $\tau$-MDP and the trajectories in the original MDP. Formally, let $\Sigma := \{ \omega \mid (s_0, s_1, \dots, s_n), s_i \in S \}$, and $\Sigma^{\tau}_{\hat{\pi}}=\{\hat{\omega} \mid \hat{\omega}\sim \mathcal{D}_{\mathcal{M}^{\tau}}^{\hat{\pi}}\}$, then define
\[
\mathcal{H}\colon \Sigma^{\tau}_{\hat{\pi}} \to  \Sigma 
\]
\[
 \hat{\omega} \mapsto (\hat{\omega}(0)(-1), \hat{\omega}(1)(-1), \dots, \hat{\omega}(n)(-1))
\]
Intuitively, this says for a trajectory $\omega$ in an original MDP, if $len(\omega)=len(\hat{\omega})$ and $\omega(i)=\hat{\omega}(i)(-1)$ for all $i\in \{0,\cdots, len(\omega)-1\}$, then $\omega$ is called the original trajectory of $\hat{\omega}$. 
According to Kolmogorov extension theorem, there exists a probability measure $\lambda_{\bar{P}^\tau}$ over $\Sigma^{\tau}_{\hat{\pi}}$, the map $\mathcal{H}$ pushes probability  $\lambda_{\bar{P}^\tau}$ into $\Sigma$, entails probability measure $\lambda_{\bar{P}^\tau}\circ \mathcal{H}^{-1}$. We denote $\mathcal{D}_{\mathcal{M}}^{\hat{\pi}}$ the probability space $(\Sigma, \lambda_{\bar{P}^\tau}\circ \mathcal{H}^{-1})$.  

Note that, the $\tau$-trajectory $\hat{\omega}$ does not conform to the format of input to STL semantics, and $\hat{\omega}$ needs to be reduced to the original trajectory for semantic interpretation.
That is, for a certain STL semantics $\tilde{\rho}$, there is $\tilde{\rho}(\hat{\omega},\varphi)=\tilde{\rho}(\omega,\varphi)$, where $\omega$ is the original trajectory of $\hat{\omega}$, and we extend the definition of $\tilde{\rho}$ over $\tau$-trajectory. 
We have the following conclusion. 

\begin{theorem}[Optimality transfer] \label{theorem_optimal}
Let $\hat{\pi}^{*}$ be any optimal policy in $\tau$-MDP. Then, executing $\hat{\pi}^{*}$ in the original MDP also maximizes the expected STL semantic value $\tilde{\rho}(\omega, \varphi)$ of the original MDP, where $\omega$ is an arbitrary trajectory generated by executing a $\tau$-MDP's policy in the original MDP, and $\varphi$ is the given STL specification.
\end{theorem}

\begin{proof}
    For a given original MDP $\mathcal{M}$ and its extended $\tau$-MDP, as well as a certain STL semantics $\tilde{\rho}$, the optimal policy $\hat{\pi}^*$ of the $\tau$-MDP is:  
    \begin{align}\label{eq_proof_1}
        \hat{\pi}^*=\mathop{\arg\max}_{\hat{\pi}}\mathop{\mathbb{E}}_{\hat{\omega}\sim \mathcal{D}_{\mathcal{M}^{\tau}}^{\hat{\pi}}}[\tilde{\rho}(\hat{\omega}, \varphi)], 
    \end{align}

    Since $\tilde{\rho}(\hat{\omega},\varphi)=\tilde{\rho}(\omega,\varphi)$ where $\omega$ is the original trajectory of $\hat{\omega}$, i.e., $len(\omega)=len(\hat{\omega})$ and $\omega(i)=\hat{\omega}(i)(-1)$ for all $i\in \{0,\cdots, len(\omega)-1\}$, we can get from Equation~(\ref{eq_proof_1}):
    \begin{align}\label{eq_proof_2}
        \hat{\pi}^* = \mathop{\arg\max}_{\hat{\pi}} \mathop{\mathbb{E}}_{\hat{\omega}\sim \mathcal{D}_{\mathcal{M}^{\tau}}^{\hat{\pi}}} [\tilde{\rho}(\omega, \varphi)].
    \end{align}

    Let $\Sigma^{\tau}_{\hat{\pi}^*}=\{\hat{\omega} \mid \hat{\omega}\sim \mathcal{D}_{\mathcal{M}^{\tau}}^{\hat{\pi}^*}\}$ be the set of all trajectories generated by $\hat{\pi}^*$ executed on $\mathcal{M}^{\tau}$, and let $\mathcal{D}_{\mathcal{M}}^{\hat{\pi}^*}$ be the induced probability space. 
    Then $\Sigma_{\hat{\pi}^*}=\{\omega \mid \omega\sim \mathcal{D}_{\mathcal{M}}^{\hat{\pi}^*}\}$ 
    denote the set of all trajectories generated by $\hat{\pi}^*$ executed on the original MDP $\mathcal{M}$. According to the definition of $\mathcal{H}$, 
    for any $\hat{\omega}\in \Sigma^{\tau}_{\hat{\pi}^*}$, there exists one and only one $\omega\in \Sigma_{\hat{\pi}^*}$ corresponding to it. 
    % such that $len(\omega)=len(\hat{\omega})$ and $\omega(i)=\hat{\omega}(i)(-1)$ for all $i\in \{0,\cdots, len(\omega)-1\}$, that is, $\Sigma_{\hat{\pi}^*}$ is also the set of original trajectories of $\Sigma^{\tau}_{\hat{\pi}^{*}}$. 
    So we can get from Equation~(\ref{eq_proof_2}):
    \begin{align}\label{eq_proof_3}
        \hat{\pi}^* = \mathop{\arg\max}_{\hat{\pi}} \mathop{\mathbb{E}}_{\omega\sim \mathcal{D}_{\mathcal{M}}^{\hat{\pi}}} [\tilde{\rho}(\omega, \varphi)].
    \end{align}
    Therefore, Theorem~\ref{theorem_optimal} holds.
    \hfill $\Box$
\end{proof}

\subsection{Causation-Guided RL Method} \label{subsec_problem}
In our online causation-guided RL method, for a given MDP model of the system $\mathcal{M}=\langle S, A, P, R \rangle$, the objective of the Problem~\ref{prob_synthesis} is to generate a good enough control policy $\pi^*$, so that the expected causation distance values of all state trajectories $\omega$ in $\mathcal{D}^{\pi^*}_{\mathcal{M}}$ with respect to specification $\varphi$ get maximized.
Based on Theorem~\ref{theorem_optimal} and \cite[Theorem 4.2]{conf/cdc/AksarayJKSB16}, the control synthesis problem in the MDP model can be transformed into the problem in the $\tau$-MDP model, i.e., the expected semantic value yielded by $\pi^*$ is bounded by that yielded by $\hat{\pi}^*$. We formalize our problem as follows.
% formally presented as follows.
\begin{problem}[Control policy synthesis problem of STL causation semantics] \label{problem_cau-synthesis}
    Let $\varphi$ be an STL specification for the system.
    Given a $\tau$-MDP model of the system $\mathcal{M}^{\tau}=\langle S^{\tau}, A, P^{\tau}, R^{\tau} \rangle$, where $P^{\tau}$ is unknown and $R^{\tau}$ is given by the online causation semantics of STL.     
    The objective is synthesizing a controller to generate a good enough control policy $\hat{\pi}^*$, so that the expected violation causation distances of all trajectories $\hat{\omega}$ in $\mathcal{D}^{\hat{\pi}^*}_{\mathcal{M}^{\tau}}$ w.r.t $\varphi$ gets maximized.
    Mathematically,
    \begin{align}\label{equ_cau-synthesic}
    \hat{\pi}^*=\mathop{\arg\max}_{\hat{\pi}}\mathop{\mathbb{E}}_{\hat{\omega}\sim \mathcal{D}_{\mathcal{M}^{\tau}}^{\hat{\pi}}}[\bar{\rho}(\hat{\omega}, \varphi)].
    \end{align}
\end{problem}
Now, we need to optimize Equation (\ref{equ_cau-synthesic}) with the RL algorithm to obtain the optimal policy. 
The implementation details and algorithm are discussed in Section~\ref{section_learning}.

\section{Learning Approach and Algorithm}\label{section_learning}
In this section, we give a concrete implementation of solving Problem~\ref{problem_cau-synthesis} with the RL algorithm, including some technical details and the setting of the reward function.

\subsection{Target Specification} \label{subsec_target_spec}
A specification of a control system is typically characterized by a combination of safety (bad things never happen) and liveness (good things eventually happen) properties~\cite{journals/dc/AlpernS87,Kindler2007}.
The safety specification is typically represented by an expression shaped like $\Box_{[0,T]}\varphi_1$, while the liveness property can be represented as $\Diamond_{[0,T]}\varphi_2$, where $T$ is a real number denoting a sufficiently large time step and also the maximum running time step of an episode of RL.
However, since real systems do not run infinitely and some scenarios require more stringent time constraints on liveness, i.e., bounded liveness~\cite{journals/dc/Lamport00}, the liveness property is denoted as $\Diamond_{[0,t]}\varphi_2$ in the STL specification, where $t\ll T$.
After time $t$, the liveness specifications will no longer impact the quantitative semantics of the entire STL specification, which will lead to a very low influence of the liveness specifications in RL (only meaningful for time $[0,t]$).

To make the liveness specification more efficiently utilized in RL, we can nest a layer of $\Box_{I}$ operators outside the liveness specification, then the liveness property becomes $\Box_{I}(\Diamond_{[0,t]}\varphi_2)$. Now, the liveness specification can be merged with the safety specification into $\Box_{I}(\Diamond_{[0,t]}\varphi_2 \wedge \varphi_1)$, which compare the form $\Box_I\varphi$, where $I=[0,T]$.
% which can then be merged with the safety specification into an STL specification of the form $\Box_I\varphi$, where $I=[0:T]$.
Thus, it is sufficient to consider only the violation causation distance $\mathscr{[R]}^{\ominus}$, i.e., $\bar{\rho}(\hat{\omega}, \varphi) = \mathscr{[R]}^{\ominus}(\hat{\omega},\varphi,0)$ in Equation (\ref{equ_cau-synthesic}).

\subsection{Sampling Window}\label{subsec_interval}
Generally, the causation distance is computed based on the signal from time $0$.
However, as the running time of the system increases, the state trajectories lengthen, reducing the efficiency of causation distance calculation. 
To address this, we compute the causation distance for partial signals only.
Inspired by the concept of the {\em horizon}~\cite{conf/rv/DokhanchiHF14} in STL, we define the {\em minimum sampling window} of the partial signal for an STL specification.
\begin{definition}[STL minimum sampling window]\label{def_horizon}
    Let $\bold{v}_{0:b}$ be a partial signal, and $\varphi$ be an STL specification. The minimum sampling window $h(\bold{v}_{0:b}, \varphi)$ is defined as:
    \begin{align*}
        &h(\bold{v}_{0:b}, \top)=  \emptyset, 
        h(\bold{v}_{0:b}, \alpha) = \emptyset, \\
        &h(\bold{v}_{0:b}, \neg\varphi) = h(\bold{v}_{0:b}, \varphi), \\
        &h(\bold{v}_{0:b}, \varphi_1\wedge\varphi_2) = h(\bold{v}_{0:b}, \varphi_1)\cup h(\bold{v}_{0:b}, \varphi_2), \\
        &h(\bold{v}_{0:b}, \varphi_1\vee\varphi_2) = h(\bold{v}_{0:b}, \varphi_1)\cup h(\bold{v}_{0:b}, \varphi_2), \\
        &h(\bold{v}_{0:b}, \Box_I\varphi) = 
            \begin{cases}
                I\cup h(\bold{v}_{0:b}, \varphi), & \text{if}\quad b>\max(I)\\
                \emptyset, & \text{otherwise}
            \end{cases}, \\
        &h(\bold{v}_{0:b}, \Diamond_I\varphi) = 
            \begin{cases}
                I\cup h(\bold{v}_{0:b}, \varphi), & \text{if}\quad b>\max(I)\\
                \emptyset, & \text{otherwise}
            \end{cases}, \\
        &h(\bold{v}_{0:b}, \varphi_1 \mathcal{U}_I \varphi_2) = &\\
        &\qquad \qquad
            \begin{cases}
                I\cup h(\bold{v}_{0:b}, \varphi_1)\cup h(\bold{v}_{0:b}, \varphi_2), & \text{if } b>\max(I)\\
                \emptyset, & \text{otherwise}
            \end{cases}. 
    \end{align*}
\end{definition}
Note that it is sufficient that the sampling window $H$ of the actual application can contain the minimum sampling window, i.e., $h(\bold{v}_{0:b}, \varphi)\subseteq H$ where $H=[0,u]$ is a positive closed interval.
The violation causation distance of the signal within the sampling window $H$ is denoted as $\bar{\rho}(\bold{v}_{b-H},\varphi)$, where $b-H$ denotes the interval $[b-u, b]$ and $\bold{v}_{b-H}$ represents the signal $\bold{v}_{b-u:b}$.
In this case, the minimum state aggregation value of the system model is 
\begin{equation}\label{equ_k}
    k = \left\lceil \frac{u}{\Delta t} \right\rceil + 1,
\end{equation}
where $\Delta t$ is the time step and $\left\lceil \cdot \right\rceil$ is the ceiling function.

\subsection{Methodology and Reward Function}
To solve Problem \ref{problem_cau-synthesis} with the RL algorithm, i.e., find the optimal policy $\hat{\pi}^{*}$, we employ the parameterized policy $\hat{\pi}_\theta(a|s^{\tau})$, which denotes the probability of selecting action $a$ given state $s^{\tau}$ with parameter $\theta$, i.e.
\begin{equation}
    \hat{\pi}_\theta(a|s^{\tau}) = \mathbb{P}[a|s^{\tau};\theta].
\end{equation}
The target function is associated with the policy parameter $\theta$ as follows:
\begin{equation}
    J(\theta)=\mathop{\mathbb{E}}_{\hat{\omega}\sim \hat{\pi}_{\theta}}[\bar{\rho}(\hat{\omega},\varphi)],
\end{equation}
where $\mathop{\mathbb{E}}\limits_{\hat{\omega}\sim \hat{\pi}_{\theta}}$ denotes the expectation operation over all traces $\hat{\omega}$ generated by using the policy $\hat{\pi}_\theta$.

Our goal is to learn a policy that maximizes the target function. 
Mathematically,
\begin{equation}
    \theta^{*} = \mathop{\arg\max}_{\theta}J(\theta).
\end{equation}
Hence, our optimal policy would be the one corresponding to $\theta^{*}$, i.e., $\hat{\pi}_{\theta}^{*}$.
We use an Actor-critic algorithm~\cite{Konda2003} to find the optimal policy.
The gradient for the Actor (policy) can be approximated as follows:
\begin{equation}\label{equ_policy-gradient}
    \nabla_{\theta}J(\theta)=\mathop{\mathbb{E}}_{\hat{\pi}_{\theta}}[\sum_{t=0}^{T-1}\nabla_{\theta}\log\hat{\pi}_{\theta}(a_t|s_t^{\tau})\cdot Q_\theta(s_t^{\tau},a_t)],
\end{equation}
where the Q-function of the actor $Q_\theta(s_t^{\tau}, a_t)$ is given as:
\begin{equation}\label{equ_Q-function}
    Q_\theta(s_t^{\tau}, a_t)=\mathop{\mathbb{E}}_{\hat{\pi}_{\theta}}[\sum_{t'=t}^{T}\bar{\rho}(s^{\tau}_{t'},\varphi)|s^{\tau}_t,a_t].
\end{equation}
Therefore, the reward function is defined as
\begin{equation}
    R^{\tau}(s^{\tau}) = \bar{\rho}(s^{\tau},\varphi).
\end{equation}
The policy gradient, given by Equation (\ref{equ_policy-gradient}), is used to find the optimal policy. The critic, on the other hand, uses another action-value function $Q^{\eta}(s^{\tau}_t, a_t)$ to estimate the Q-function $Q_\theta(s^{\tau}_t, a_t)$ of the actor.

\subsection{Smooth Approximation of Online Causation Semantics} \label{subsec_smooth}

The Q-value significantly influences the types of policies that can be expressed and regions that can be explored in RL.
In fact, smoother Q-values play a crucial role in RL~\cite{conf/icml/Nachum0TS18} as they lead to smoother policies~\cite{conf/icml/ShenLJWZ20}.
However, the current Q-value in Equation (\ref{equ_Q-function}) lacks smoothness due to the non-smooth reward function $\bar{\rho}(s^{\tau}_{t'},\varphi)$.
For instance, consider the example in Figure~\ref{fig_QCauM}, the violation causation distance curves exhibit the characteristics of a piecewise function, which is non-smooth at segmentation points.
This lack of smoothness primarily arises from the $\max$ and $\min$ functions within the semantics, where an increase (or decrease) of a single value does not affect the result of the $\max$ and $\min$ functions, unless it is the minimum (or maximum) value.

Therefore, we use the log-sum-exp~\cite{conf/Chen12} approximation to smooth the $\max$ and $\min$ functions in the causation semantics, i.e.,
\begin{align}
    \max(x_1,\cdots,x_n)\sim \frac{1}{\beta}\log\sum_{i=1}^{n}e^{\beta x_i},
\end{align}
\begin{align}
    \min(x_1,\cdots,x_n)\sim -\frac{1}{\beta}\log\sum_{i=1}^{n}e^{-\beta x_i},
\end{align}
where $\beta>0$ is a user-defined parameter and
\begin{align}
\begin{split}
    \max(x_1,\cdots,x_n) & \leq \frac{1}{\beta}\log\sum_{i=1}^{n}e^{\beta x_i} \\ & \leq  \max(x_1,\cdots,x_n) + \frac{1}{\beta}\log n,
\end{split}
\end{align}
\begin{align}
\begin{split}
    \min(x_1,\cdots,x_n) & \geq -\frac{1}{\beta}\log\sum_{i=1}^{n}e^{-\beta x_i} \\ & \geq \min(x_1,\cdots,x_n) - \frac{1}{\beta}\log n.
\end{split}
\end{align}
Note that $\frac{1}{\beta}\log\sum_{i=1}^{n}e^{\beta x_i}$ and $-\frac{1}{\beta}\log\sum_{i=1}^{n}e^{-\beta x_i}$ can approximate the $\max$ and $\min$ function with arbitrary accuracy by selecting a large $\beta$. 
However, a larger $\beta$ value will not always be better, as a large $\beta$ may cause the approximation function to lose differentiability.

\begin{algorithm}[h]
\caption{Online Causation-Guided RL Algorithm}
\label{alg_causation}
\begin{algorithmic}[1]
  \Require 
    Actor parameters $\theta$, critic parameters $\eta$; 
    learning rates $\alpha_\theta$ (actor) and $\alpha_\eta$ (critic); 
    discount factor $0 \le \gamma < 1$;
    sampling length $k$.
  \For{each episode}
    \State Initialize state $s_0$, $s_0^\tau=(s_0)^{k}$;\label{alg_initial_begin}
    \State Current state $s^{\tau}_t = s^{\tau}_0$;\label{alg_initial_end}
    \For{$t = 0,1,2,\dots$ until termination}
      \State Sample action $a_t \sim \hat{\pi}_\theta(\cdot \mid s^{\tau}_t)$;\label{alg_action}
      \State Execute $a_t$ and observe next state $s^{\tau}_{t+1}$;\label{alg_excute}
      \State Compute reward $r^\tau=R^{\tau}(s^{\tau}_{t+1})$;\label{alg_reward}
      \Statex \makebox[\linewidth][r]{\textcolor{gray}{\textit{\# $R^{\tau}$ is the smooth causation semantics}}}
      \State Sample the next action: $\displaystyle a' = \hat{\pi}_\theta(\cdot \mid s^{\tau}_{t+1})$; \label{alg_next-state}
      \State Compute temporal-difference error for action-value at time $t$: 
        \[
          \quad \delta_t \;\gets\; r^{\tau}_t + \gamma\,Q_\eta(s^\tau_{t+1}, a') \;-\; Q_\eta(s^\tau_t, a_t);
        \]
      \State Update the parameters of the critic: 
        \[
           \eta \;\gets\; \eta + \alpha_\eta\,\delta_t\,\nabla_\eta Q_\eta(s^\tau_t, a_t);
        \]
      \State Update the parameters of the actor: \label{alg_update}
        \[
          \quad \theta \;\gets\; \theta + \alpha_\theta\,\nabla_\theta Q_\eta(s^\tau_t, \hat{\pi}_\theta(a_t \mid s^\tau_t));
        \]
      \State Update state $s^\tau_t \gets s^\tau_{t+1}$
    \EndFor
  \EndFor
\end{algorithmic}
\end{algorithm}

\subsection{Online Causation-Guided RL Algorithm}\label{subsec_algorithm}

The solution to Problem~\ref{problem_cau-synthesis} can be summarized as Algorithm~\ref{alg_causation}.
A run of a system from start to end is called an {\em episode}.
The inputs to the algorithm are the randomly initialized actor parameter $\theta$ and critic parameter $\eta$, the learning rates $\alpha_\theta$ for actor and $\alpha_\eta$ for critic, the discount factor $\gamma$, and the sampling length $k$ (calculated by Equation (\ref{equ_k})). 

We explain the steps of the algorithm as follows:
\begin{itemize}
    \item[1)] For each episode, initialize the state $s^{\tau}_0$ of $\tau$-MDP (introduced in Section~\ref{subsec_system_model}) and assign it to the current state $s^\tau_t$ (Line~\ref{alg_initial_begin}-\ref{alg_initial_end}).  
    \item[2)] For each step in an episode, first select the current action $a_t$ according to the current policy $\hat{\pi}_{\theta}$, after which execute the action $a_t$ and observe the next state $s^{\tau}_{t+1}$ (Line~\ref{alg_action}-\ref{alg_excute}).
    \item[3)] Then the reward for the next state is computed. Note that the reward function here is a smooth approximation of the STL online causation semantics (detailed in Section~\ref{subsec_smooth}), taking the causation distance of the next $\tau$-state (a $k$-step state sequence) as the reward (Line~\ref{alg_reward}).
    \item[4)] Finally, select the next action $a_{t+1}$ according to the current policy $\hat{\pi}_{\theta}$, compute the temporal difference error, and update the network parameters of the critic and the actor (Line~\ref{alg_next-state}-\ref{alg_update}).
\end{itemize}
The algorithm will end after multiple episodes, and the resulting $\hat{\pi}_\theta$ is the synthesized control policy.
Compared to regular actor-critic algorithms, the main differences in online causality-guided RL algorithms are that 1) the inputs to the actor and critic networks are the states of the $\tau$-MDP, i.e., a sequence of states rather than a single state, and 2) the reward function is given by a smooth approximation of the STL causation semantics.

%% file: Chapter/4-experiments.tex
\section{Experiments} \label{sec_experiment}
We have implemented a prototype tool for our method and conducted experiments to evaluate its effectiveness and compare it with other STL-guided RL methods. 
All experiments are performed on an Ubuntu 22.04 machine equipped with an Intel Core i7-2700F CPU, an NVIDIA GeForce RTX 3080 GPU, and 32GB of RAM. The evaluated artifact and the experimental data are available at \url{https://doi.org/10.5281/zenodo.16879140}.
% can be found in the supplemental materials.

\subsection{Benchmarks}

\begin{figure}[t]
    \centering
    \subcaptionbox{Cart-Pole}{
        \includegraphics[width=0.45\linewidth]{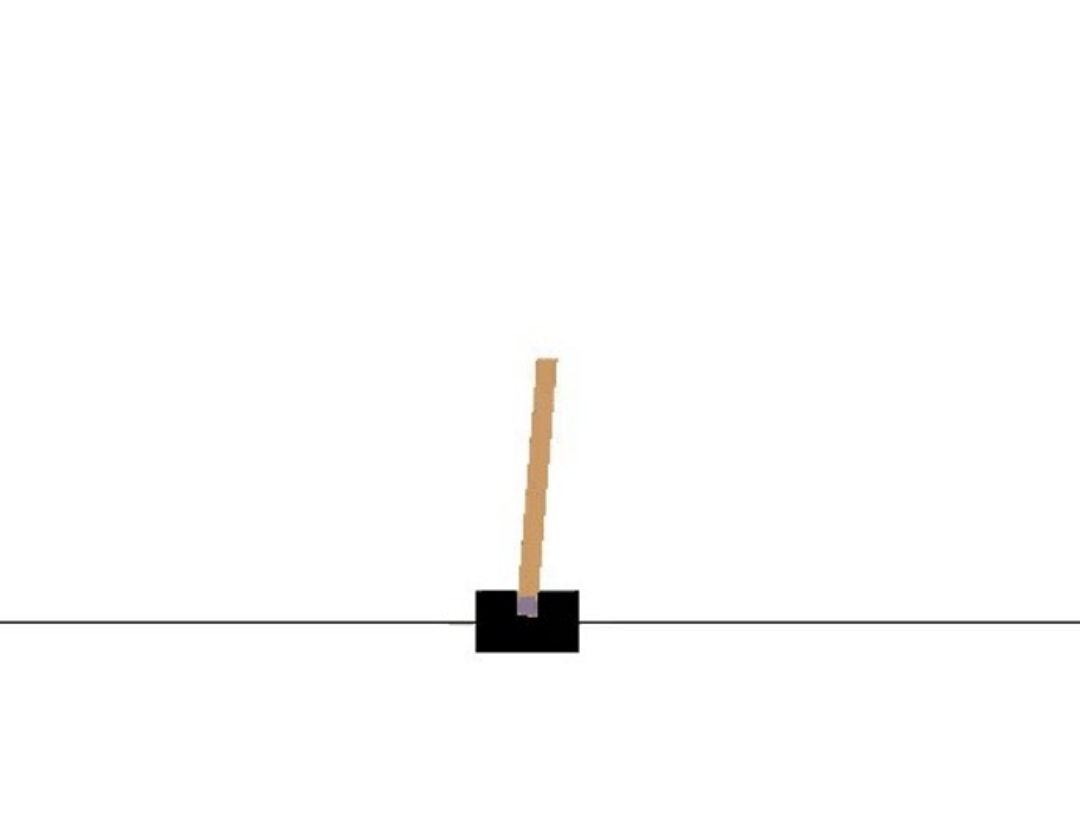}
        \label{fig_env_cartpole}
    }   
    \subcaptionbox{Reach-Avoid}{
        \includegraphics[width=0.45\linewidth]{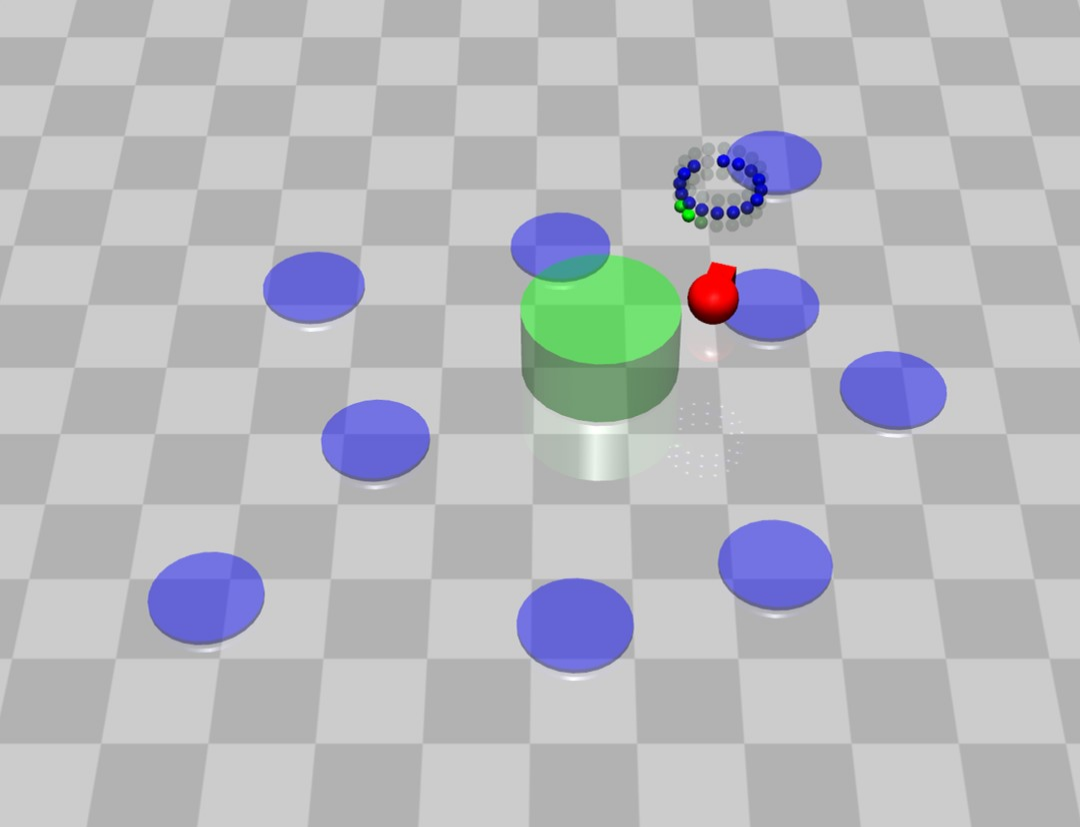}
        \label{fig_env_pointgoal}
    }
    \subcaptionbox{Hopper}{
        \includegraphics[width=0.45\linewidth, trim=0 0 0 235, clip]{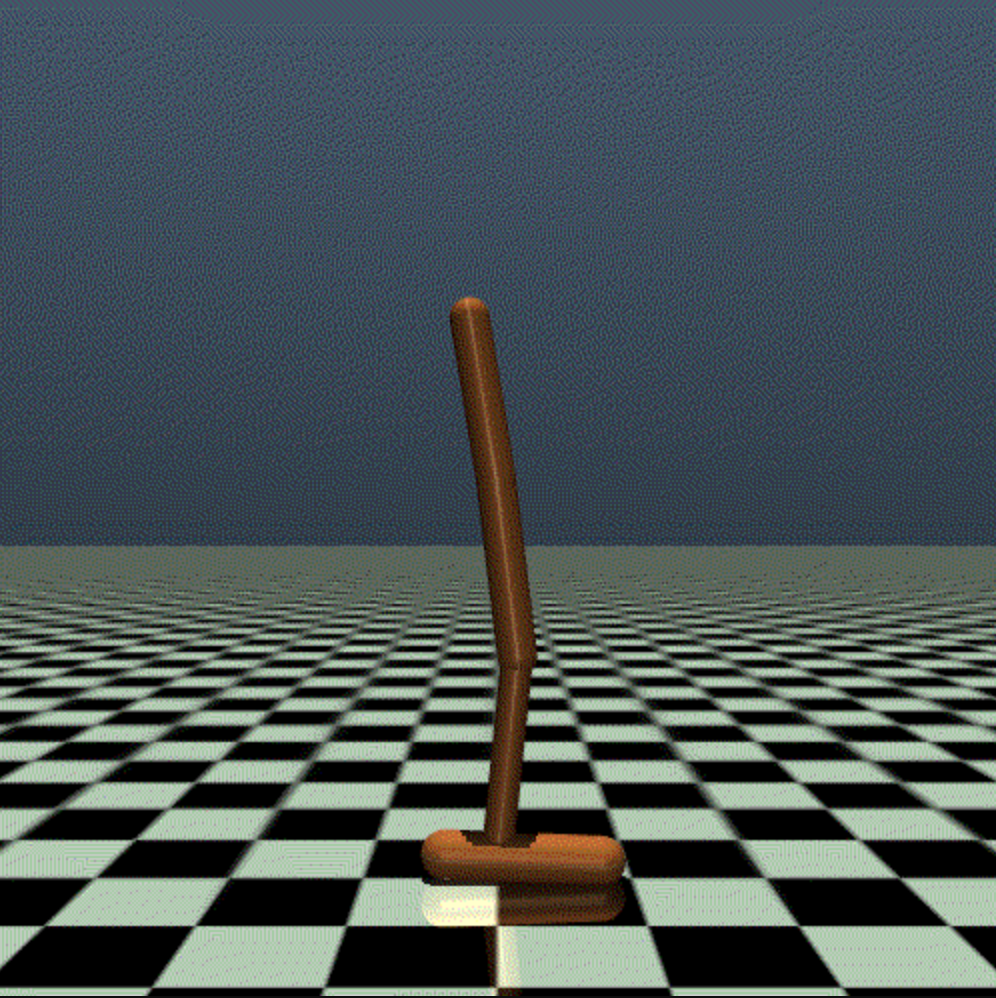}
        \label{fig_env_hopper}
    }
    \subcaptionbox{Walker}{
        \includegraphics[width=0.45\linewidth]{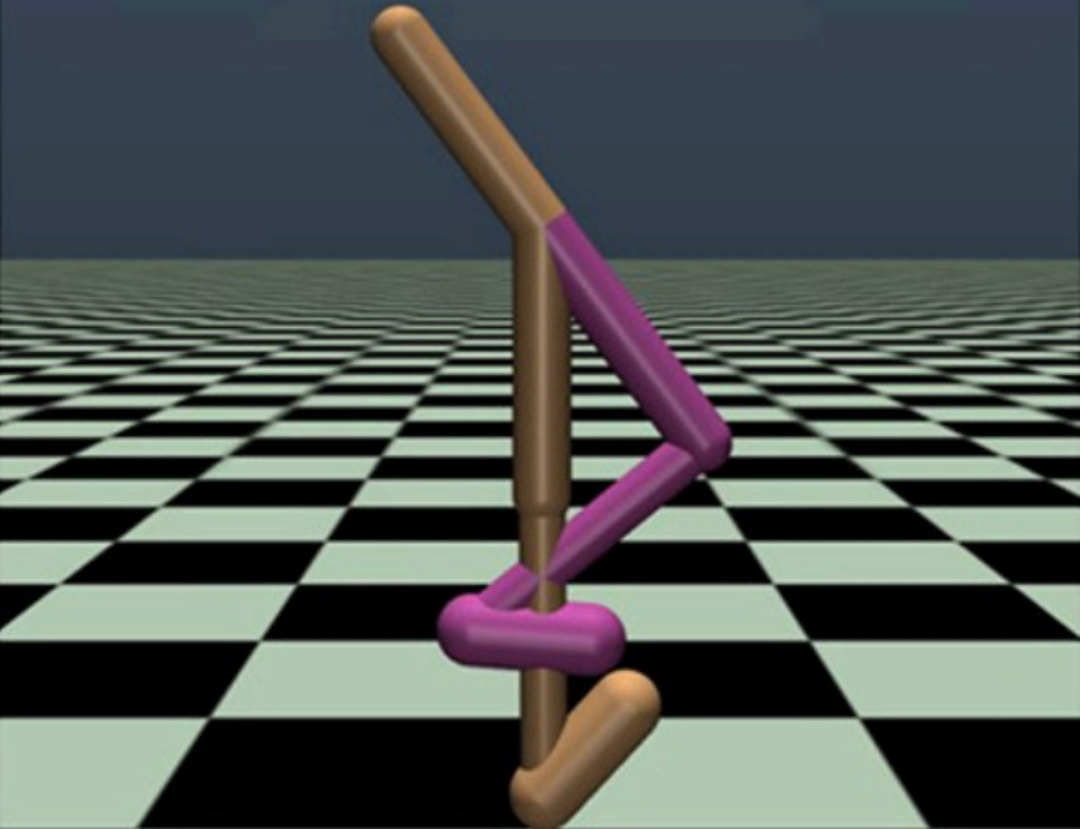}
        \label{fig_env_walker}
    }
    \caption{Benchmarks used in experiments.}
    \label{fig_envs}
\end{figure}
We utilize the OpenAI Gym platform~\cite{ref_journals/corr/BrockmanCPSSTZ16} as simulators for CPSs.
The experiments were conducted on four benchmarks shown in Figure~\ref{fig_envs}: Cart-Pole, Reach-Avoid, Hopper, and Walker.
Each benchmark, along with its corresponding target STL specification, is described below.
\paragraph{Cart-Pole}
The first benchmark is the classical Cart-Pole balancing problem described in~\cite{journals/tsmc/BartoSA83}, where a cart is controlled using a simple bang-bang controller that pushes it left or right. 
The state space of the environment is $(x, \dot{x}, \theta, \dot{\theta})$, where $x$ is the displacement of the cart from the origin, $\dot{x}$ is the speed of the cart, $\theta$ is the angle of displacement of the pole from the upright position, and $\dot{\theta}$ is the angular velocity of the pole rotation.
The original target is to balance a pole on the cart for as long as possible.
Our experiments have a more stringent target that can be formulated as the following STL specification:
\begin{equation*}
    \varphi=\Box_{[0,\infty]}(\Diamond_{[0,10]}(|\dot{x}|<0.1)\wedge(|x|<0.5)\wedge(|\theta|<0.1)),
\end{equation*}
which means the cart has to  achieve a position at $(-0.5,0.5)$, a velocity less than $0.1$, and an angle less than $0.1$ in $10$ steps for any moment.

\paragraph{Reach-Avoid}
The Reach-Avoid benchmark presents a typical reach-avoid problem on the Safety Gym platform~\cite{Achiam2019BenchmarkingSE}, where a point navigates to a green goal while avoiding contact with the 9 unsafe hazards on the map. 
The point has 44 states, including 16 states that represent Lidars that detect the distance and direction to the goal, and another 16 states measure the distance to hazards.
The target of the experiment is that the point has to keep a safe distance from the hazards all the time and reach the goal range in $40$ steps.
The target can be formulated into the following STL specification:
\begin{equation*}
    \varphi=\Box_{[0,\infty]}(\Diamond_{[0,40]}(d_g\leq r_g)\wedge(d_c>r_c))
\end{equation*}
where $d_g$ is the distance to the goal, $d_c$ is the distance to the closest hazard, and $r_g=0.3$ and $r_c=0.2$ are the specified hazard radius and goal radius respectively.
Both $d_c$ and $d_g$ are calculated from Lidar readings in the states.

\paragraph{Hopper}
The third benchmark is derived from Mujoco environments~\cite{conf/iros/TodorovET12}.
The Hopper is a two-dimensional one-legged figure. 
The goal is to make hops that move in the forward (right) direction by applying torques on the three hinges connecting the four body parts.
The state space of Hopper is an $11$-dimensional array, including positional values, velocities, angles and angle velocities of different body parts.
If the height of the top part is greater than $0.7$ and the angle is within the range $[-1,1]$, then Hopper's posture is considered healthy.
The target of the experiment is that Hopper has to maintain a healthy posture all the time and reach a speed of $0.5$ in $15$ steps.
The target can be formulated into the following STL specification:
% \vspace{2mm}
\begin{equation*}
    \varphi=\Box_{[0,\infty]}(\Diamond_{[0,15]}(v_x>0.5)\wedge(z>0.7)\wedge(|a|<1))
\end{equation*}
where $v_x$ is the velocity of the top in the x-direction, $z$ is the z-coordinate of the top and $a$ is the angle of the top.

\paragraph{Walker}
The last benchmark is also derived from Mujoco environments. 
The Walker is a two-dimensional two-legged figure.
The goal is to coordinate both sets of feet, legs, and thighs to move in the forward (right) direction by applying torques on the six hinges connecting the six body parts.
The Walker has 17 states, including positional values, velocities, angles and angle velocities of different body parts.
If the height of the top part is within the range $[0.8, 2]$ and the angle is within the range $[-1,1]$, then Walker's posture is considered healthy.
The target of the experiment is that the Walker has to maintain a healthy posture all the time and reach a speed of $0.5$ in $15$ steps.
The target can be formulated into the following STL specification:
\begin{equation*}
    \varphi=\Box_{[0,\infty]}(\Diamond_{[0,15]}(v_x>0.5)\wedge(0.8<z<2)\wedge(|a|<1))
\end{equation*}
where $v_x$ is the velocity of the top in the x-direction, $z$ is the z-coordinate of the top and $a$ is the angle of the top.

\subsection{Experimental Setup}
\paragraph{Training setting} 
We have implemented our causation-guided RL method using Stable-Baseline3~\cite{journals/jmlr/RaffinHGKED21}, a framework developed and maintained by OpenAI.
Like other STL semantics, causation semantics can be combined with various RL algorithms. 
We follow the choices commonly adopted in the benchmarks~\cite{conf/aaai/0004S23,Achiam2019BenchmarkingSE,journals/tcad/JiangLK24}, using the Proximal Policy Optimization (PPO) algorithm~\cite{journals/corr/SchulmanWDRK17} to train the control policy for the Cart-Pole and Reach-Avoid benchmarks, while the state-of-the-art Actor-Critic (SAC) algorithm~\cite{conf/icml/HaarnojaZAL18} is applied to the Hopper and Walker benchmarks.
% The Proximal Policy Optimization (PPO) algorithm~\cite{journals/corr/SchulmanWDRK17} is used to train the control policy for the Cart-Pole and Reach-Avoid benchmarks, while the state-of-the-art Actor-Critic (SAC) algorithm~\cite{conf/icml/HaarnojaZAL18} is applied to the Hopper and Walker benchmarks.
We integrate the smooth causation-STL semantics on top of the online monitoring tool Breach~\cite{conf/cav/Donze10}.
The parameter $\beta$ of the log-sum-exp function for the STL causation semantics is set to $10$ and the length of partial signals for each benchmark is calculated by Equation (\ref{equ_k}). 
Table~\ref{table_train_setting} presents the maximum steps per episode, training episodes, and the structure of neural networks.

\begin{table}[t]
% \vspace{1mm}
\caption{Settings of training parameters ($m_{epi}$: max step per episode, $n_{total}$: the number of total steps, $lr$: the learning rate, and $stru_{NN}$: the structure of the hidden layers of the neural network, i.e., the number of nodes per layer).}
\label{table_train_setting}
\begin{center}
\tabcolsep=0.04\linewidth
\begin{tabular}{lrrrr}
\toprule
Benchmark & $m_{epi}$ & $n_{epi}$& $lr$& $stru_{NN}$\\
\midrule
\textbf{Cart-Pole}  & 500  & 500  &4e-4 & [64, 64]\\
\textbf{Reach-Avoid}& 1000 & 500   &1e-5 & [1024, 512, 512]\\                 
\textbf{Hopper}     & 1000 & 1600  &3e-4 &[64, 64]\\
\textbf{Walker}     & 1000 & 1600  &3e-4 &[64, 64]\\
\bottomrule
\end{tabular}
\end{center}
\end{table}

\paragraph{Evaluation settings} 
In this paper, we compare our causation-guided RL method (\textbf{CAU}) with the baseline RL method using the hand-crafted reward function (\textbf{BAS})~\cite{journals/jmlr/RaffinHGKED21} and three most-related STL-guided RL methods, i.e., using classic robust semantics directly (\textbf{CLS})~\cite{conf/cav/Donze10,journals/fmsd/DeshmukhDGJJS17}, log-sum-exp approximation of STL robust semantics (\textbf{LSE})~\cite{conf/amcc/LiMB18} and Smoothened-Summation-Subtraction approximation of STL robust semantics (\textbf{SSS})~\cite{conf/aaai/0004S23}.
These types of semantics have also been implemented on top of Breach~\cite{conf/cav/Donze10}. 

Let $T$ denote the length of a single episode, and $\varphi$ denote the given STL specification.
We use several metrics to measure the performance of each method:
\begin{itemize}
    \item[-]  \emph{Success or Failure}  ($\mathsf{S/F}$)~\cite{Achiam2019BenchmarkingSE}. If the training results converge and achieve the training goal (note that strict satisfaction of the given specification is not required), it is considered a success $\mathsf{S}$; otherwise, it is considered a failure $\mathsf{F}$.
    \item[-]  \emph{Full Satisfaction} ($\mathsf{Full\mbox{-}SAT}$). For the full specification $\varphi$, the $\mathsf{Full\mbox{-}SAT}$ for a whole episode of length $T$ is defined as $\mathsf{Full\mbox{-}SAT}=\mathbb{I}[\rho(s_{0:T}, \varphi, 0)]$, where $\mathbb{I}$ is an indicator function that equals $1$ when the full specification $\varphi$ is satisfied, i.e., $\rho(s_{0:T}, \varphi, 0)\geq 0$.
    \item[-]  \emph{Safety Satisfaction} ($\mathsf{Safety\mbox{-}SAT}$). Let $\varphi_s$ denote the safety component of the specification $\varphi$. The $\mathsf{Safety\mbox{-}SAT}$ for a full episode of length $T$ is defined as $\mathsf{Safety\mbox{-}SAT}=\mathbb{I}[\rho(s_{0:T}, \varphi_s, 0)]$, where $\mathbb{I}$ is an indicator function that equals $1$ when the safety component $\varphi_s$ is satisfied, i.e., $\rho(s_{0:T}, \varphi_s, 0)\geq 0$.    
    \item[-]  \emph{Cost Return} ($\mathsf{CR}$)~\cite{Achiam2019BenchmarkingSE}. The average cost return for $N$ episodes is defined as $\mathsf{CR} =\frac{1}{N} \sum_{i=0}^{N} \sum_{t=0}^{T}c_t $, where $c_t$ is fixed for each step of safety violation in the environment.
    This metric can be interpreted as the average policy's performance in terms of safety violations. Therefore, the lower, the better.
\end{itemize}
Note that for the comparison of these methods, we first compare their $\mathsf{S/F}$. For the successful methods, we then compare their $\mathsf{Full\mbox{-}SAT}$, $\mathsf{Safety\mbox{-}SAT}$ and $\mathsf{CR}$.

\subsection{Experimental Results}

\begin{figure}[!t]
    \centering    
    \captionsetup[subfigure]{skip=0pt}
    \subcaptionbox{Cart-Pole\label{fig_compare_cartpole}}{
        \includegraphics[width=0.444\linewidth]{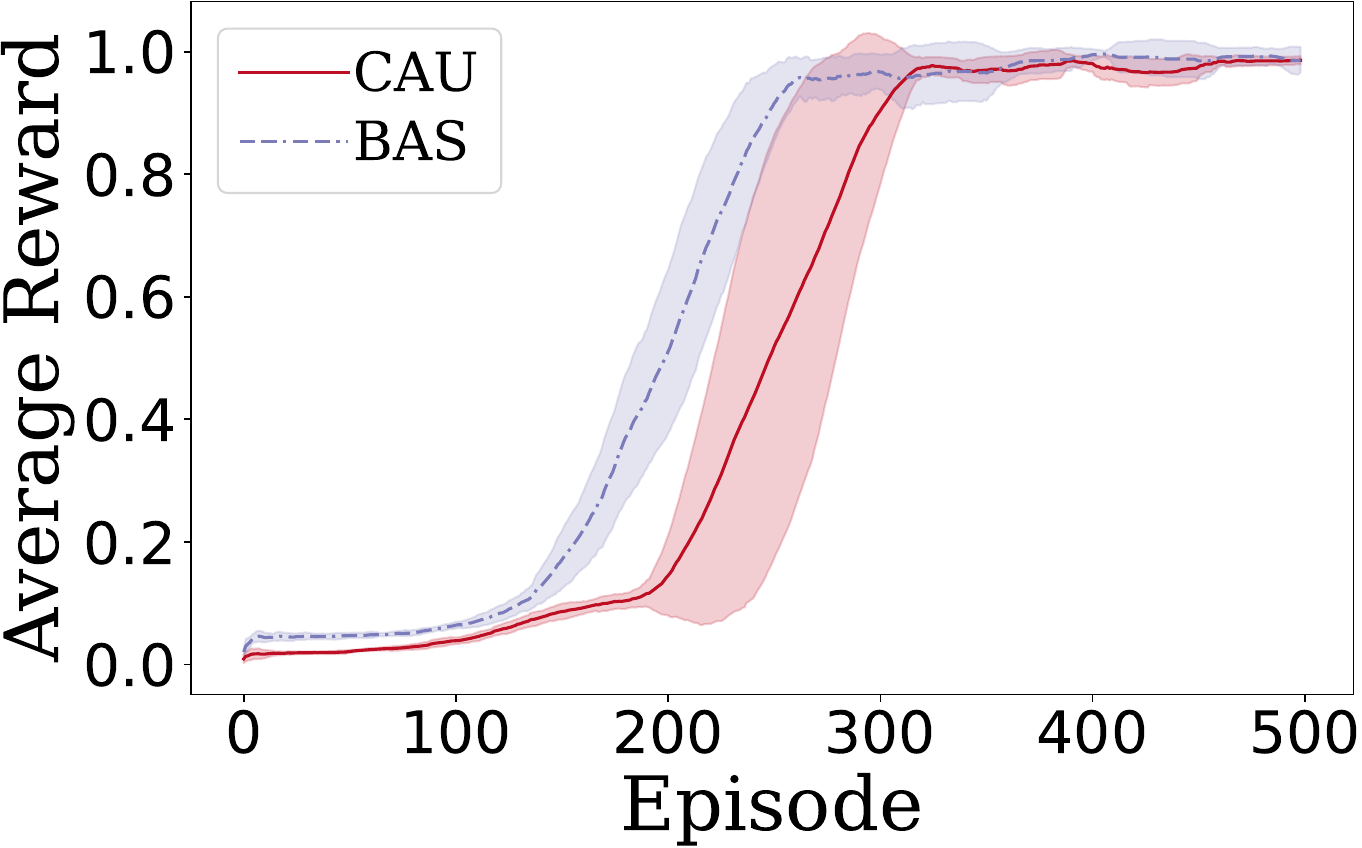}        
    }
    \subcaptionbox{Reach-Avoid\label{fig_compare_reachavoid}}{
        \includegraphics[width=0.452\linewidth]{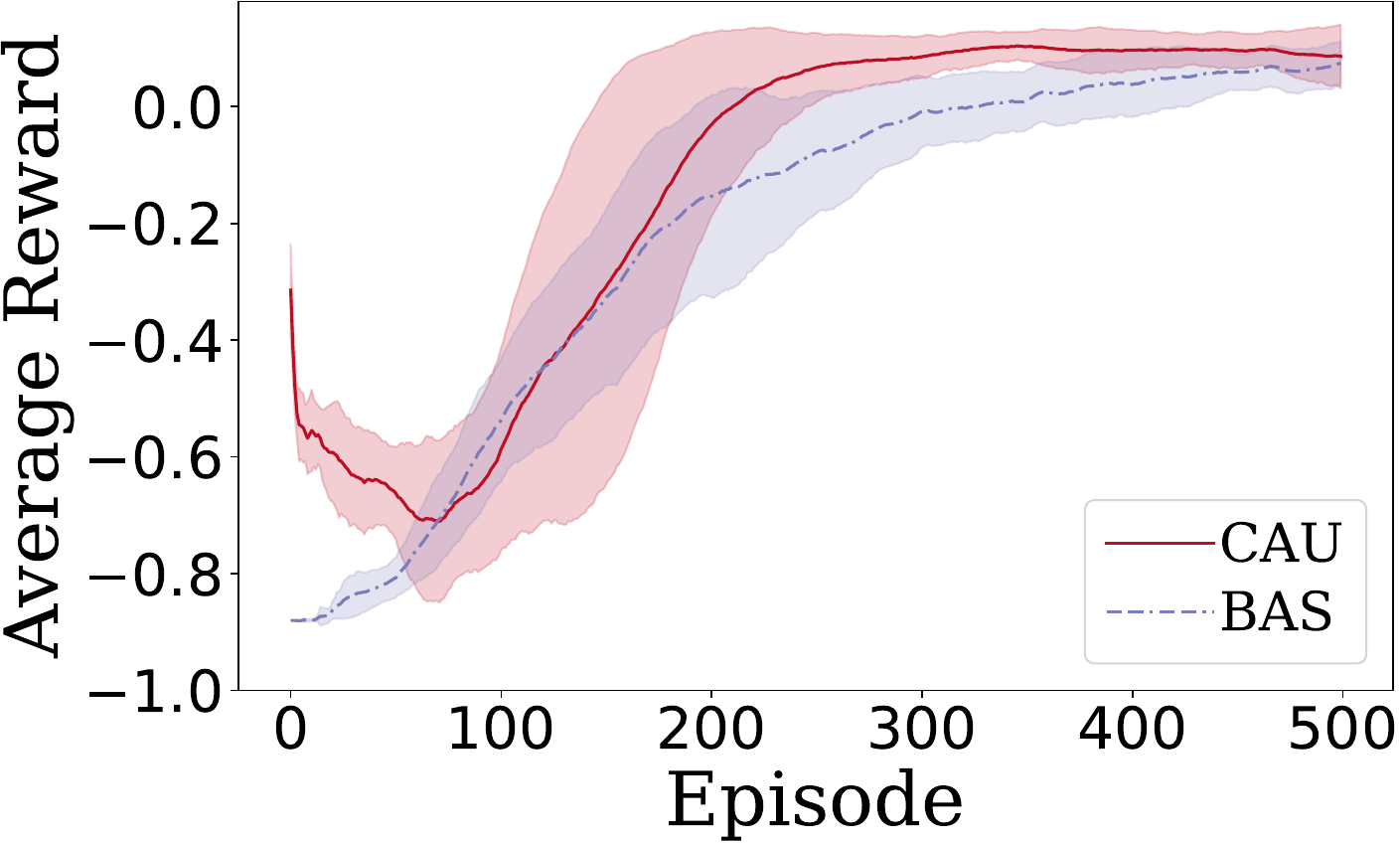}        
    }
    \subcaptionbox{Hopper\label{fig_compare_hopper}}{
        \includegraphics[width=0.444\linewidth]{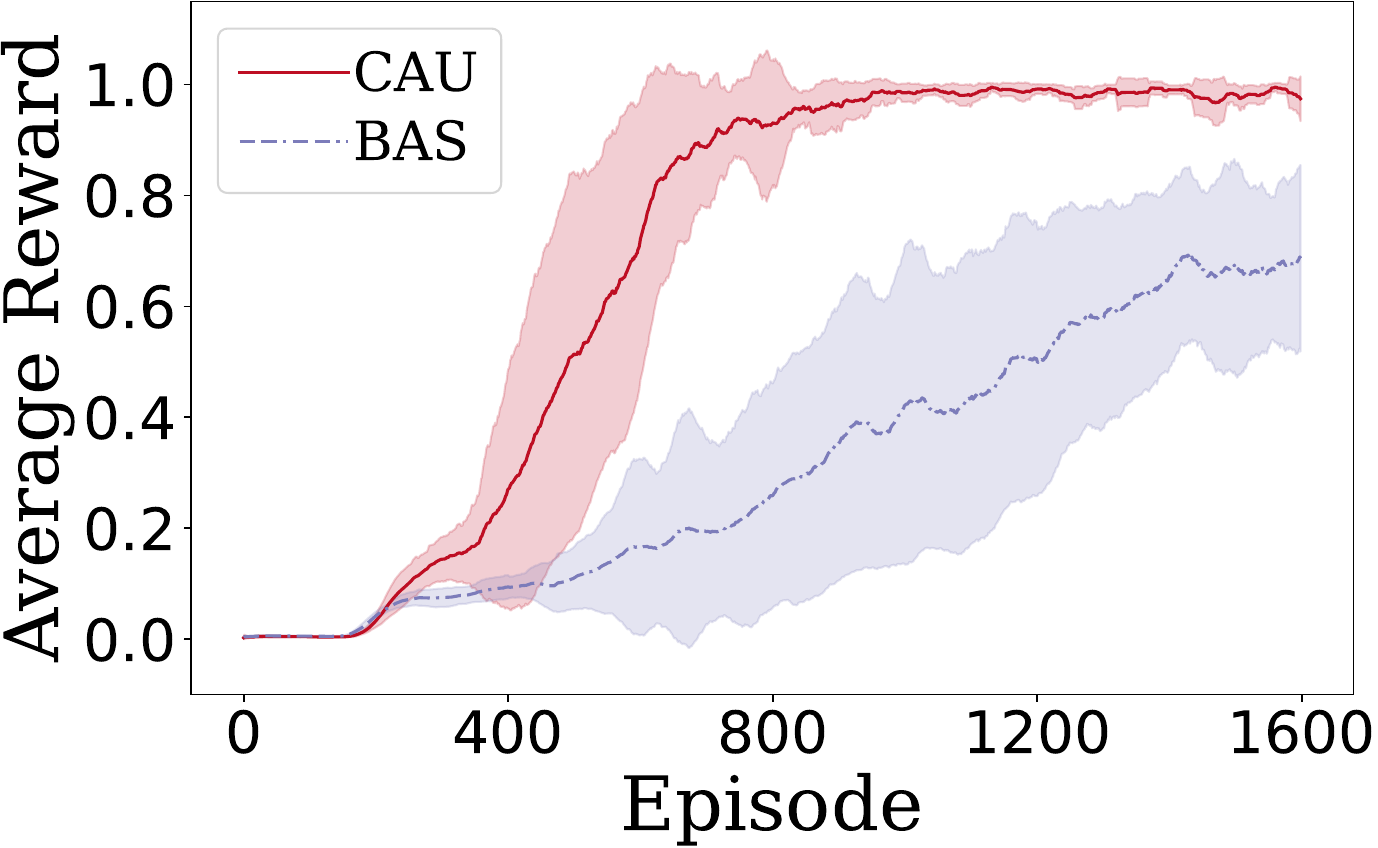} 
    }
    \subcaptionbox{Walker\label{fig_compare_walker}}{
        \includegraphics[width=0.444\linewidth]{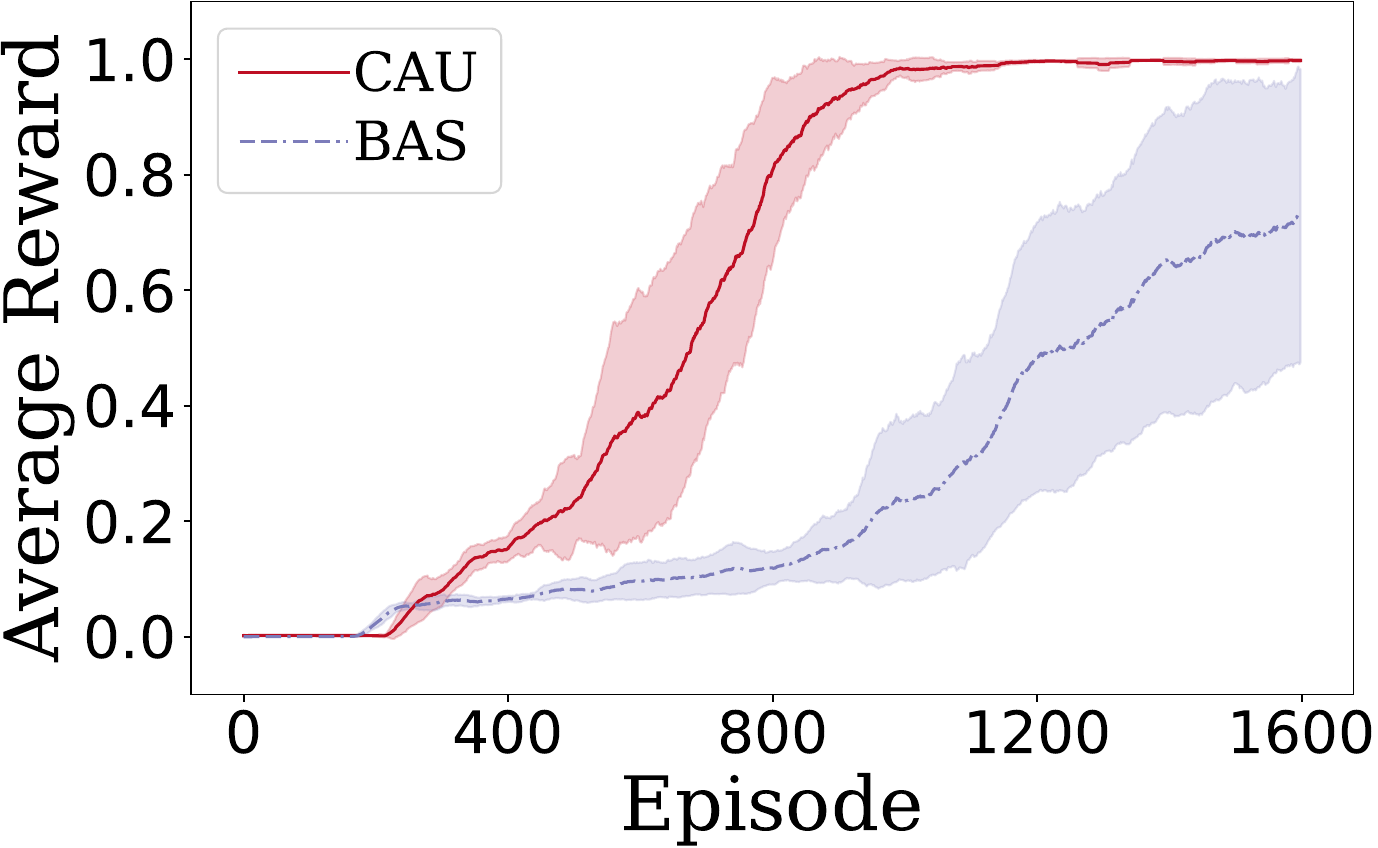}     
    }
    \caption{Learning curves for training processes of \textbf{CAU} and \textbf{BAS} method.}
    
    \label{fig_results_base}
\end{figure}

\begin{figure}[!t]
    \centering
    \captionsetup[subfigure]{skip=0pt}
    \subcaptionbox{Cart-Pole\label{fig_result_cartpole}}{
        \includegraphics[width=0.457\linewidth]{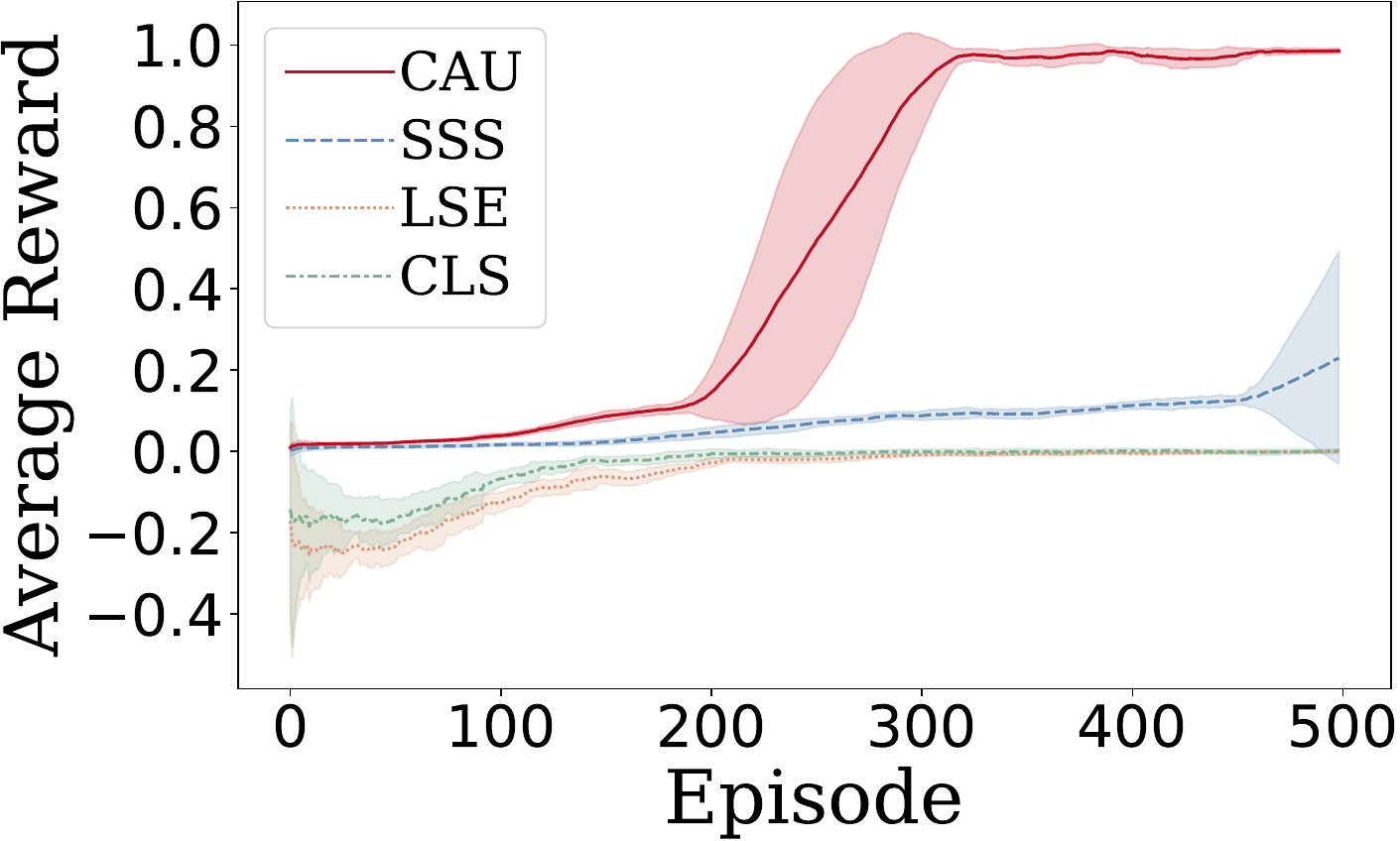}        
    }
    \subcaptionbox{Reach-Avoid\label{fig_result_reachavoid}}{
        \includegraphics[width=0.457\linewidth]{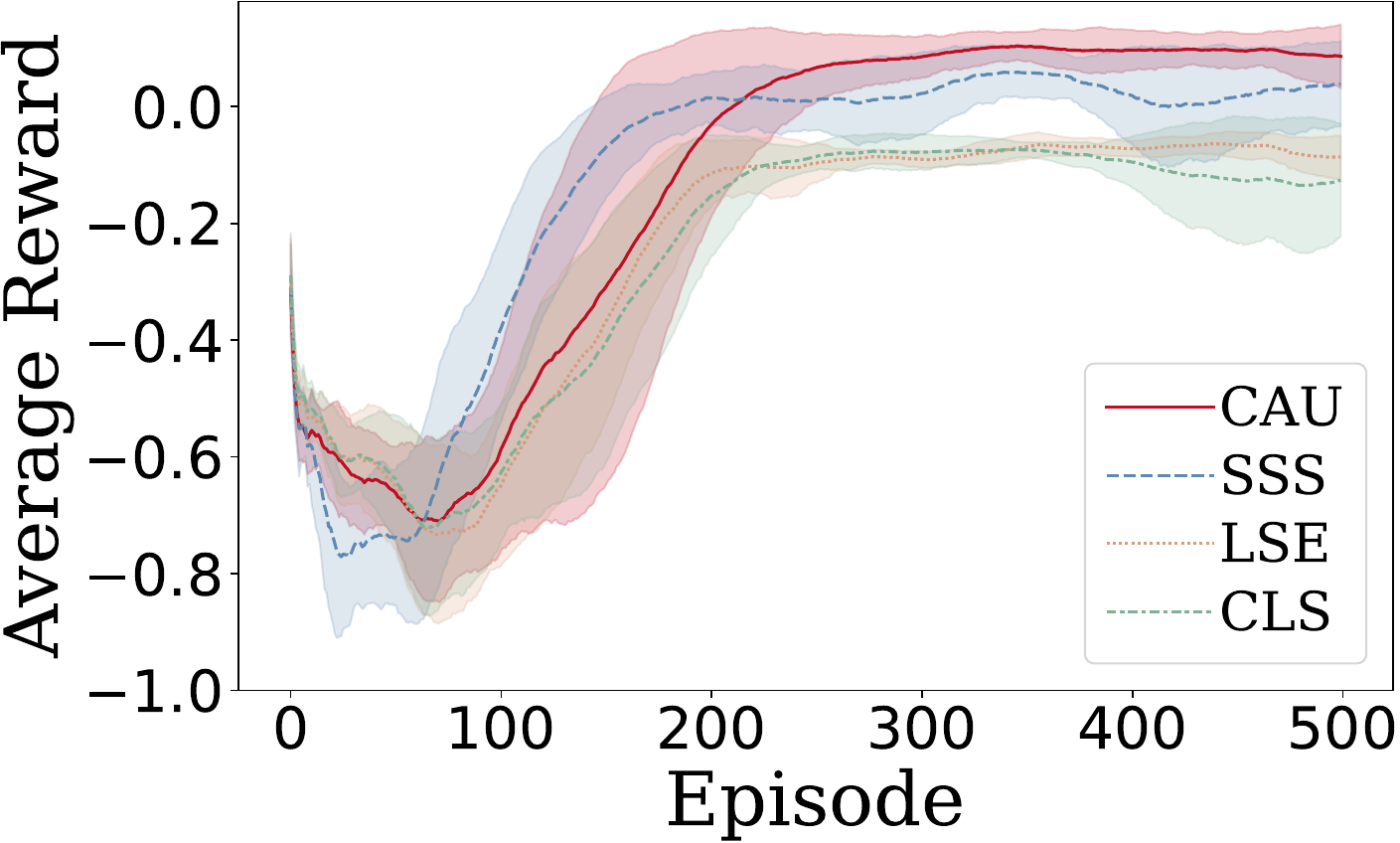}        
    }
    \subcaptionbox{Hopper\label{fig_result_hopper}}{
        \includegraphics[width=0.457\linewidth]{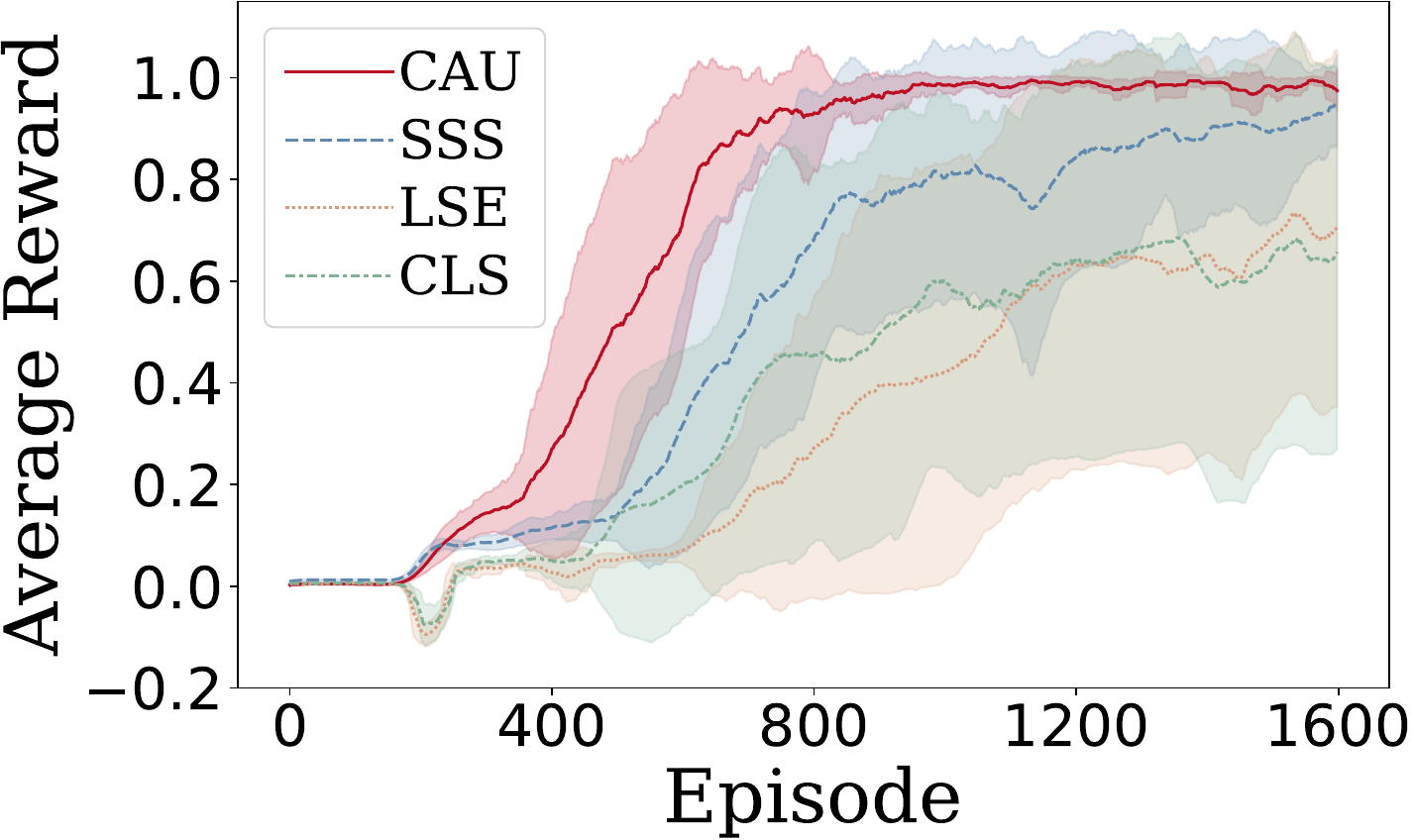} 
    }
    \subcaptionbox{Walker\label{fig_result_walker}}{
        \includegraphics[width=0.457\linewidth]{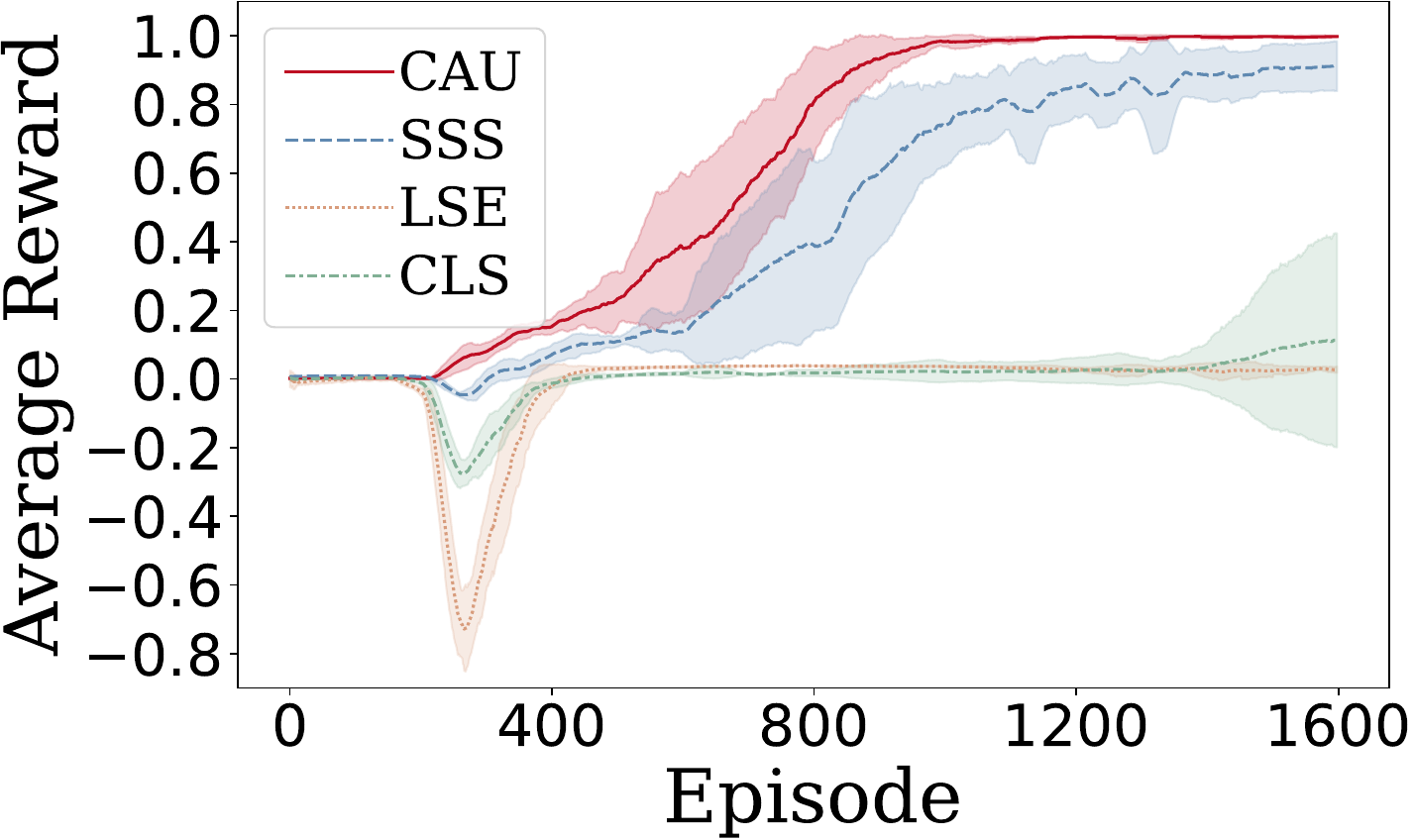}     
    }
    \caption{Learning curves for training processes of \textbf{CAU} and other STL-guided method.}
    
    \label{fig_results_stl}
\end{figure}

\begin{table*}[t]
% \vspace{2mm}
\caption{Experimental results of different methods under different environments.}
\label{table_evaluation}
\begin{center}
\resizebox{0.98\textwidth}{!}{
\begin{tabular*}{\hsize}{@{}@{\extracolsep{\fill}}lccrrr@{}}
\toprule
Benchmark & Method & $\mathsf{S/F}$ & $\mathsf{Full\mbox{-}SAT}$& $\mathsf{Safety\mbox{-}SAT}$ & $\mathsf{CR}$  \\
\midrule
\multirow{5}{*}{\textbf{Cart-Pole}} 
              & \textbf{CAU}  & $\mathsf{S}$   & \hlt 0.4290$\pm$0.3522 & \hlt 0.8520$\pm$0.1481 & \hlt 0.0031$\pm$0.0033\\ 
              & \textbf{SSS}  & {\color{gray}$\mathsf{F}$}&{\color{gray}-}&{\color{gray}-}&{\color{gray}-} \\
              & \textbf{LSE}  & {\color{gray}$\mathsf{F}$}&{\color{gray}-}&{\color{gray}-}&{\color{gray}-} \\
              & \textbf{CLS}  & {\color{gray}$\mathsf{F}$}&{\color{gray}-}&{\color{gray}-}&{\color{gray}-} \\
              & \textbf{BAS}  & $\mathsf{S}$   &  0.1020$\pm$0.2439 & 0.5430$\pm$0.3958 &  0.1536$\pm$0.2537\\ 
\midrule 
\multirow{5}{*}{\textbf{Reach-Avoid}} 
              & \textbf{CAU}  & $\mathsf{S}$  &0.0$\pm$0.0 & \hlt 0.4220$\pm$0.0890  & \hlt 0.0167$\pm$0.0153\\ 
              & \textbf{SSS}  & $\mathsf{S}$  &0.0$\pm$0.0 &0.3680$\pm$0.1487  &0.0298$\pm$0.0289\\
              & \textbf{LSE}  & $\mathsf{S}$  &0.0$\pm$0.0 &0.3280$\pm$0.1025  &0.0503$\pm$0.0390\\
              & \textbf{CLS}  & $\mathsf{S}$  &0.0$\pm$0.0 &0.2320$\pm$0.1655  &0.0533$\pm$0.0423\\
              & \textbf{BAS}  & $\mathsf{S}$  &0.0$\pm$0.0 &0.4540$\pm$0.0486 &0.0010$\pm$0.0019 \\
\midrule 
\multirow{5}{*}{\textbf{Hopper}} 
              & \textbf{CAU}  & $\mathsf{S}$ & \hlt 0.8270$\pm$0.2709 & \hlt 0.9520$\pm$0.1282  & \hlt 5.8e-6$\pm$1.57e-4\\ 
              & \textbf{SSS}  & $\mathsf{S}$ &0.6670$\pm$0.4246 &0.8990$\pm$0.2997  &0.0011$\pm$0.0034\\              
              & \textbf{LSE}  & $\mathsf{S}$ &0.7780$\pm$0.3896 &0.7780$\pm$0.3896  &0.0014$\pm$0.0028\\
              & \textbf{CLS}  & $\mathsf{S}$ &0.6200$\pm$0.4376 &0.6200$\pm$0.4376  &0.0022$\pm$0.0030\\
              & \textbf{BAS}  & $\mathsf{S}$   &  0.3540$\pm$0.3791 & 0.5540$\pm$0.4073 &  0.0050$\pm$0.0130\\
\midrule                  
\multirow{5}{*}{\textbf{Walker}} 
              & \textbf{CAU}  & $\mathsf{S}$  & \hlt 0.7120$\pm$0.2787 & \hlt 1.0$\pm$0.0     & \hlt 0.0$\pm$0.0  \\ 
              & \textbf{SSS}  & $\mathsf{S}$  &0.3560$\pm$0.3821 &0.9960$\pm$0.0066  &4e-6$\pm$7e-6\\
              & \textbf{LSE}  & {\color{gray}$\mathsf{F}$}&{\color{gray}-}&{\color{gray}-}&{\color{gray}-}\\
              & \textbf{CLS}  & {\color{gray}$\mathsf{F}$}&{\color{gray}-}&{\color{gray}-}&{\color{gray}-}\\
              & \textbf{BAS}  & $\mathsf{S}$   &  0.2090$\pm$0.2678 & 0.7570$\pm$0.3542 & 0.0005$\pm$ 0.0010\\
\bottomrule
\end{tabular*}   
}
\end{center}
\end{table*}

For each benchmark, we perform training over $10$ random seeds, and count the average rewards per $50$ rounds for each training.
Figure~\ref{fig_results_base} and Figure~\ref{fig_results_stl} show the comparison of our method with baseline and our method with other STL-guided RL methods in the training process, respectively, where the folds are the mean values of the 10 trainings, and the shaded areas indicate the standard deviation.
Due to the varying reward scales provided by different STL semantics, we normalized the training results for each experiment.
As illustrated by learning curves, our \textbf{CAU} method achieves favorable training outcomes across different benchmarks. 
Additionally, we have tested the synthesized controller on 100 distinct seeds to evaluate the training effectiveness of the methods.
The test results are shown in Table~\ref{table_evaluation}.
It is intuitively obvious from the learning curves and evaluation table that \textbf{CAU}, \textbf{SSS} and \textbf{BSA} are clearly superior to the other two methods.
In order to compare the three methods more objectively, we conducted statistical tests, and the results are shown in Table~\ref {table_testing}.
We adopt the Mann-Whitney U test~\cite{book/mann1947test} with $\alpha = 0.05$ as the confidence value of the null hypothesis of no significant difference; in case of a significant difference, we use Vargha and Delaney’s $\hat{A}_{12}$ effect size~\cite{journals/vargha2000critique} to assess the degree of significance. 

From the training process and test results, we can observe that \textbf{CAU} outperforms other methods:
\begin{itemize}
    \item[-] In the Cart-Pole benchmark, only our \textbf{CAU} is able to converge (as shown in Figure~\ref{fig_result_cartpole}), while in this case, none of the other three STL-based methods could be trained successfully.
    Also, although \textbf{BAS} converges slightly faster than \textbf{CAU}, the evaluation is much worse for $\mathsf{Full\mbox{-}SAT}$, $\mathsf{Safety\mbox{-}SAT}$ and $\mathsf{CR}$.
    \item[-] In the Reach-Avoid benchmark, every STL-guided method converges and returns a controller.
    Although \textbf{CAU} converges more slowly than \textbf{SSS}, it has a more stable training effect, which is clearly shown in Figure~\ref{fig_result_reachavoid}.
    In addition, \textbf{CAU} has also been evaluated to have higher safety assurance (higher $\mathsf{Safety\mbox{-}SAT}$ and lower $\mathsf{CR}$) compared to other STL-based methods.
    The difference in evaluation values between \textbf{CAU} and \textbf{BAS} is also statistically negligible.
    The reason for $\mathsf{Full\mbox{-}SAT}$ being $0$ in the evaluation results is that the initial state of the Point is too far from the target to reach it within 40 time-steps (it was tested to be reachable within about 200 time-steps).
    \item[-] In the Hopper benchmark, all STL-guided methods successfully trained controllers.
    However, \textbf{CAU} converges fastest and most stably as shown in Figure~\ref{fig_result_hopper}, and the obtained controller achieves higher $\mathsf{Full\mbox{-}SAT}$ and $\mathsf{Safety\mbox{-}SAT}$, and lower $\mathsf{CR}$.
    \item[-] In the Walker benchmark, both \textbf{CAU} and \textbf{SSS} can successfully converge, while \textbf{BAS} has not yet completely converged. 
    However, \textbf{CAU} converges faster and more stable than \textbf{SSS} as shown in Figure~\ref{fig_result_walker}, and achieves perfect $\mathsf{Full\mbox{-}SAT}$, much higher $\mathsf{Safety\mbox{-}SAT}$, and zero $\mathsf{CR}$. 
\end{itemize}

Overall, compared to the other three STL-based methods, our method has a 24\%-100\% improvement in $\mathsf{Full\mbox{-}SAT}$, a 0.4\%-14.7\% improvement in $\mathsf{Safety\mbox{-}SAT}$, and a 44\%-100\% decrease in $\mathsf{CR}$.

\begin{table}[t]
\caption{Statistical testing of comparing \textbf{CAU} with \textbf{BAS} and \textbf{SSS}.}
\label{table_testing}
\begin{center}
\begin{threeparttable}
\begin{tabular*}{\hsize}{@{}@{\extracolsep{\fill}}llcc@{}}
\toprule
 &  & \textbf{BAS} & \textbf{SSS}  \\
\midrule
\multirow{3}{*}{\textbf{Cart-Pole}}  
            & $\mathsf{Full\mbox{-}SAT}$  & \ding{51}\ding{51}\ding{51}  & \ding{51}\ding{51}\ding{51} \\ 
            & $\mathsf{Safety\mbox{-}SAT}$  & \ding{51}\ding{51}  & \ding{51}\ding{51}\ding{51} \\
            & $\mathsf{CR}$  & \ding{51}\ding{51}  & \ding{51}\ding{51}\ding{51} \\
\midrule
\multirow{3}{*}{\textbf{Reach-Avoid}}  
            & $\mathsf{Full\mbox{-}SAT}$  & -  & - \\ 
            & $\mathsf{Safety\mbox{-}SAT}$  & -  & \ding{51} \\
            & $\mathsf{CR}$  & -  & \ding{51}\ding{51}\ding{51} \\
\midrule 
\multirow{3}{*}{\textbf{Hopper}} 
            & $\mathsf{Full\mbox{-}SAT}$  & \ding{51}\ding{51}\ding{51}  & \ding{51}\\ 
            & $\mathsf{Safety\mbox{-}SAT}$  & \ding{51}\ding{51}\ding{51} & -\\
            & $\mathsf{CR}$  & \ding{51}\ding{51}\ding{51}  & - \\
\midrule                  
\multirow{3}{*}{\textbf{Walker}} 
            & $\mathsf{Full\mbox{-}SAT}$  & \ding{51}\ding{51}\ding{51} & \ding{51}\ding{51}\\ 
            & $\mathsf{Safety\mbox{-}SAT}$  & \ding{51}\ding{51}\ding{51} &\ding{51}\ding{51} \\
            & $\mathsf{CR}$  & \ding{51}\ding{51}\ding{51}  & \ding{51}\ding{51} \\
\bottomrule

\end{tabular*}  
\begin{tablenotes} 
\item  negligible: -\;\;	small better.:\ding{51} \;\;	medium better.: \ding{51}\ding{51} \;\;  large better.: \ding{51}\ding{51}\ding{51}
\end{tablenotes}
\end{threeparttable}
\end{center}
\end{table}

\subsection{Discussion}

% 分析可扩展性：
% 1）motitor的消耗：causation计算的讨论
% 2）tau-MDP的消耗

% 参数的影响：
% 不同k的实验

% \begin{table}[t]
% \caption{The average training time for per-step of different methods.}
% \label{table_train_time}
% \begin{center}
% \begin{threeparttable}
% \begin{tabular*}{\hsize}{@{}@{\extracolsep{\fill}}lcr@{}}
% \toprule
% Benchmark & Method & time(s) \\
% \midrule
% \multirow{4}{*}{\textbf{Cart-Pole}} 
%               & \textbf{CAU}  &  0.001996$\pm$1.19e-5\\ 
%               & \textbf{SSS}  &  0.001394$\pm$1.85e-5\\
%               & \textbf{LSE}  &  0.001174$\pm$9.76e-5\\
%               & \textbf{CLS}  &  0.001061$\pm$1.27e-5\\
% \midrule 
% \multirow{4}{*}{\textbf{Reach-Avoid}} 
%               & \textbf{CAU}  &  0.006040$\pm$2.21e-4\\ 
%               & \textbf{SSS}  &  0.004140$\pm$1.63e-5\\
%               & \textbf{LSE}  &  0.003616$\pm$5.69e-6\\
%               & \textbf{CLS}  &  0.004056$\pm$1.73e-5\\
% \midrule 
% \multirow{4}{*}{\textbf{Hopper}} 
%               & \textbf{CAU}  &  0.007829$\pm$2.83e-5\\ 
%               & \textbf{SSS}  &  0.005666$\pm$4.61e-5\\              
%               & \textbf{LSE}  &  0.005353$\pm$4.52e-5\\
%               & \textbf{CLS}  &  0.005743$\pm$2.08e-4\\
% \midrule                  
% \multirow{4}{*}{\textbf{Walker}} 
%               & \textbf{CAU}  &  0.007656$\pm$6.56e-5\\ 
%               & \textbf{SSS}  &  0.005487$\pm$6.74e-5\\
%               & \textbf{LSE}  &  0.005856$\pm$1.96e-4\\
%               & \textbf{CLS}  &  0.005864$\pm$2.15e-4\\
% \bottomrule
% \end{tabular*}  
% \end{threeparttable}
% \end{center}
% \end{table}

\begin{table}[t]
\caption{The average training time for per-step of different methods. Time is reported in seconds.}
\label{table_train_time}
\begin{center}
\begin{threeparttable}
\begin{tabular*}{\hsize}{@{}@{\extracolsep{\fill}}lcccc@{}}
\toprule
Benchmark & \textbf{CAU}& \textbf{SSS}& \textbf{LSE}& \textbf{CLS}\\
\midrule
\textbf{Cart-Pole} & 0.001996& 0.001394& 0.001174& 0.001061\\
\midrule 
\textbf{Reach-Avoid} & 0.006040& 0.004140& 0.003616& 0.004056\\
\midrule 
\textbf{Hopper} & 0.007829& 0.005666& 0.005353& 0.005743\\
\midrule                  
\textbf{Walker} & 0.007656& 0.005487& 0.005856& 0.005864\\ 
\bottomrule
\end{tabular*}  
\end{threeparttable}
\end{center}
\end{table}

\begin{table*}[t]
% \vspace{2mm}
\caption{Experimental results and average training time for per-step of different $k$.}
\label{table_tau}
\begin{center}
\begin{tabular*}{\hsize}{@{}@{\extracolsep{\fill}}lcrrrr@{}}
\toprule
Benchmark & $k$  & $\mathsf{Full\mbox{-}SAT}$& $\mathsf{Safety\mbox{-}SAT}$& $\mathsf{CR}$& per-step time (s)\\
\midrule
\multirow{5}{*}{\textbf{Cart-Pole}} 
              & $11$  &0.4290$\pm$0.3522& 0.8520$\pm$0.1481& 3.104e-3$\pm$3.306e-3& 0.001996\\ 
              & $16$  &0.2700$\pm$0.1647& 0.6880$\pm$0.3212& 1.281e-2$\pm$2.347e-2& 0.002387\\
              & $21$  &0.1570$\pm$0.2514& 0.4150$\pm$0.2777& 1.678e-2$\pm$1.412e-2& 0.003056\\
              & $26$  &0.1970$\pm$0.1818& 0.6790$\pm$0.2238& 7.399e-3$\pm$8.810e-3& 0.003854\\
              & $31$  &0.1540$\pm$0.1973& 0.5350$\pm$0.3065& 2.328e-2$\pm$3.412e-2& 0.004785\\
\midrule 
\multirow{5}{*}{\textbf{Hopper}} 
              & $16$  &0.8270$\pm$0.2709& 0.9520$\pm$0.1282& 5.770e-6$\pm$1.568e-4& 0.007829\\ 
              & $21$  &0.9000$\pm$0.1522& 1.0$\pm$0.0& 0.0$\pm$0.0& 0.008406\\
              & $26$  &0.8020$\pm$0.3176& 1.0$\pm$0.0& 0.0$\pm$0.0& 0.009134\\
              & $31$  &0.8120$\pm$0.3420& 0.9970$\pm$0.0064& 3.012e-6$\pm$6.432e-6& 0.001005\\
              & $36$  &0.8740$\pm$0.1581& 0.9980$\pm$0.0040& 2.001e-6$\pm$4.010e-6& 0.001120\\
\bottomrule
\end{tabular*}   
\end{center}
\end{table*}

% \begin{table}[t]
% \vspace{2mm}
% \caption{The average training time for per-step of different $k$.}
% \label{table_tau_time}
% \begin{center}
% \begin{tabular*}{\hsize}{@{}@{\extracolsep{\fill}}lrrrr@{}}
% \toprule
% \multirow{2}{*}{\textbf{Cart-Pole}} 
%               & $k=11:$ 0.001996& $k=16:$ 0.002387& $k=21:$ 0.003056\\ 
%               & $k=26:$ 0.003854& $k=31:$ 0.004785&\\
% \midrule 
% \multirow{2}{*}{\textbf{Hopper}} 
%               & $k=16:$0.001996& $k=21:$0.002387& $k=26:$0.003056\\ 
%               & $k=31:$0.003854& $k=36:$0.004785&\\
% \bottomrule
% \end{tabular*}   
% \end{center}
% \end{table}

\paragraph{Scalability}
% \subsubsection{Scalability}
Compared with other STL-guided RL methods, concerns about the additional overhead of \textbf{CAU} mainly involve two aspects:
1) whether computing the causation semantics introduces additional computational cost; and 
2) whether the use of $\tau$-MDP leads to increased overhead. 
We analyze the scalability of the \textbf{CAU} method from these two aspects.
\begin{itemize}
    \item [1)] We adopt the latest implementation of \textbf{CAU}~\cite{conf/fm/ZhangAAH24}, which has shown that computing causation semantics takes the same magnitude of time as computing robust semantics, and therefore it does not introduce much overhead in comparison to classical robust semantics. % （待补充详细信息）
    \item [2)]  Although introducing the $\tau$-MDP increases the computational cost of the neural networks (due to higher input dimension and larger replay buffer), the overhead is limited and only depends on the parameter $k$, which corresponds to the time bound of the liveness property—typically a small integer. Table~\ref{table_train_time} summarizes the average per-step training time of \textbf{CAU} and other STL-guided RL methods. Although the use of $\tau$-MDP increases the state dimensionality by a factor of $6$ to $41$ in different benchmarks, the per-step training time remains within the same order of magnitude.
\end{itemize}

\paragraph{Parameter}
We further analyzed the impact of the parameter $k$ on training performance and computational cost in the Cart-Pole and Hopper benchmarks.  
Based on the minimum sampling window, we conducted four additional experiments using larger sampling windows in increments of $5$ time steps, and compiled the evaluation results and average single-step training time (as shown in Table~\ref{table_tau}).
The results indicate that
\begin{itemize}
    \item [1)] increasing $k$ does not significantly improve training performance, and overly long historical state sequences may even introduce noise and hinder policy synthesis,
    \item [2)] while training time increases sub-linearly with $k$ (the power exponent is $0.85$ for Cart-Pole and $0.43$ for Hopper).
\end{itemize}
In general, the minimum sampling window can be used as the initial setting for $k$, while the optimal trade-off between computational cost and training performance may require case-by-case analysis.

%% file: Chapter/5-related_work.tex
\section{Related Work} \label{sec_relatedwork}

Model-free RL~\cite{journals/tnn/SuttonB98} has become a popular method for synthesizing controllers for unknown dynamic systems.
In RL, the policy of an {\em agent} is trained based on feedback from a reward function, aiming to maximize cumulative reward. 
Therefore, a poorly designed reward function can lead to the development of policies that do not align with the intended objectives.
In some RL-based controller synthesis approaches~\cite{journals/corr/LillicrapHPHETS15,conf/icml/LevineK13,journals/ftrob/DeisenrothNP13,conf/aaai/FultonP18,journals/ml/HafnerR11}, the reward function is designed by a control engineer who has complete knowledge of the system dynamics. 
However, this approach is often case-specific, and designing an appropriate reward function becomes increasingly challenging as the system's dimensionality and nonlinearity increase.
Moreover, in safety-critical systems, the agent may adopt policies that maximize the expected total rewards but produce unsafe or impractical operations; this phenomenon is known as {\em reward hacking}~\cite{conf/nips/Hadfield-Menell17,journals/apin/YuanYGDL19}.

Recent developments have applied temporal logic~\cite{conf/focs/Pnueli77} to reward generation, guiding RL synthesis to obtain control policies that satisfy formal specifications.
For instance, {\em linear temporal logic} (LTL)-guided model-free RL algorithms have been designed to maximize the probability of satisfying the formal specifications~\cite{conf/cdc/HasanbeigKAKPL19, conf/icra/Bozkurt0ZP20, journals/ral/CaiHXAK21, conf/icra/CuiZLY23}.
However, reward signals derived from LTL~\cite{books/daglib/0016866} specifications are typically sparse, which hinders the effectiveness of RL in complex tasks.
Li et al.~\cite{conf/iros/LiVB17, conf/amcc/LiMB18} proposed {\em truncated linear temporal logic} (TLTL) with quantitative semantics and used its robustness degree as a reward function for RL, which improves the learning speed.
Nevertheless, LTL still lacks the ability to describe real-time constraints.

In addition, plenty of works have explored the use of STL semantics to define reward functions for RL tasks. 
Aksaray et al.~\cite{conf/cdc/AksarayJKSB16} applied STL robust semantics to reward generation in Q-learning.
Kalagarla et al.~\cite{conf/cdc/Kalagarla0N21} proposed an STL-guided RL method that considers performance constraints and two-layer nested temporal operators.
However, these methods are based on offline semantics and lack extensions in deep-RL.
Balakrishnan and Deshmukh proposed using STL robust semantics of partial signals to design a reward function for deep RL~\cite{conf/iros/0001D19}, but the robust semantics are still not smooth. 
Singh and Saha~\cite{conf/aaai/0004S23} proposed a smooth robust semantic over partial signals, showing good performance in guiding RL.

\begin{table}[t]
\vspace{1.8mm}
\caption{Comparison with other RL reward generation methods}
\label{table_reward_method}
\begin{center}
\setlength{\tabcolsep}{3pt}
\scalebox{0.92}{
\begin{tabular}{lcccc}
\toprule
 & \textbf{Automation} & \textbf{Expressiveness}  & \textbf{Quantification}  & \textbf{Informativeness} \\
\midrule
Hand-crafted  &\ding{55} & \ding{51} &\ding{51} &\ding{51} \\
LTL   &\ding{51}\ding{51}\ding{51}  &\ding{51}\ding{51}   &\ding{55} &\ding{51} \\
TLTL    &\ding{51}\ding{51}\ding{51} &\ding{51}\ding{51} &\ding{51} &\ding{51} \\
STL-robust  &\ding{51}\ding{51}\ding{51} &\ding{51}\ding{51}\ding{51} &\ding{51} &\ding{51}\ding{51} \\
Ours     &\ding{51}\ding{51}\ding{51}  &\ding{51}\ding{51}\ding{51}  &\ding{51} &\ding{51}\ding{51}\ding{51} \\
\bottomrule
\end{tabular}}
\end{center}
\vspace{-1.8mm}
\end{table}

We list above reward assignment methods and our method with 5 attributes in TABLE~\ref{table_reward_method}, in which \ding{55} indicates non-conformity of the attributes and \ding{51} and its number indicates conformity of the attributes.

\begin{itemize}
    \item[-] Compared with baseline RL using \emph{hand-crafted} rewards, the other 4 category methods are \emph{formal specification}-based, which allows engineers to easily specify system requirements by formal languages; these formally-specified requirements can be \textbf{automatically} parsed and translated to rewards, saving significant efforts of hand-crafting rewards in a case-by-case manner.
    \item[-] Compared with \emph{LTL}-based rewards, besides STL's advantages in \textbf{expressiveness}~\cite{conf/formats/MalerN04}, STL-based methods (last 2 columns) enable \textbf{quantification} of rewards, thereby exhibiting distinct strengths in \textbf{informativeness}. 
    \item[-] Other STL-guided RL methods still suffer from the information masking problem due to the monotonicity of STL robust semantics.
    Instead, our method leverages the online causation semantics of STL, which reflect the instantaneous status of system evolution more faithfully than robust semantics, and thus enjoy the benefits leading to more accurate rewards.
\end{itemize}

%% file: Chapter/6-conclusion.tex
\section{Conclusion} \label{sec_conclusion}
We propose a novel causation-guided RL approach for controller synthesis of STL. The experimental results demonstrate that our approach outperforms the existing relevant STL-guided RL methods across multiple complex benchmarks. 
Although our method incurs a small amount of additional overhead, it is acceptable compared to the improvement in satisfaction of goal.
% For future work, we aim to investigate more effective reward generation methods based on STL causation semantics and apply these techniques to practical cyber-physical systems, such as autonomous vehicles.
For future work, we aim to investigate reward generation methods with lower computational overhead and higher goal satisfaction based on STL causation semantics, and to apply these techniques to practical cyber-physical systems such as autonomous vehicles.

%% file: Chapter/8-acknowledgements.tex
\section*{Acknowledgments}
We thank the anonymous reviewers for their valuable comments, which significantly improved the quality of this paper. We also thank Shenghua Feng for his help in improving this paper. Xiaochen Tang and Miaomiao Zhang are supported by the Natural Science Foundation of China (NSFC) Grant No.~62472316 and No.~62032019. Zhenya Zhang is supported by the JSPS Grant No.~JP25K21179 and JST BOOST Grant No.~JPMJBY24D7. Jie An is supported by the National Key R\&D Program of China under grants No.~2022YFA1005100, No.~2022YFA1005101 and No.~2022YFA1005103, the NSFC grants No.~W2511064, No.~62192732 and No.~62032024, and the ISCAS Basic Research Grant No.~ISCAS-JCZD202406.